\documentclass[journal=dce]{CUP-JNL-DTM}%

\usepackage{graphicx}
\usepackage{multicol,multirow}
\usepackage{amsmath,amssymb,amsfonts}
\usepackage{mathrsfs}
\usepackage{amsthm}
\usepackage{rotating}
\usepackage{appendix}
\usepackage{ifpdf}
\usepackage[T1]{fontenc}
\usepackage{newtxtext}
\usepackage{newtxmath}
\usepackage{textcomp}
\usepackage{xcolor}
\usepackage{lipsum}
\usepackage[colorlinks,allcolors=blue]{hyperref}
\usepackage{subfigure}
\usepackage{booktabs}
\usepackage{graphicx}
\usepackage{multirow}
\usepackage{rotating}
\usepackage{enumitem}
\newcommand{\cm}{\checkmark} % Checkmark
\newcommand{\xm}{\times}     % Cross
\usepackage{tikz}

\theoremstyle{definition}

\numberwithin{equation}{section}

\jname{Data/Math}
\articletype{ARTICLE TYPE}
\jyear{YEAR}
\begin{document}

\begin{Frontmatter}

\title[Article Title]{Can Time-Series Foundation Models Perform Building Energy Management Tasks?}

% There is no need to include ORCID IDs in your .pdf; this information is captured by the submission portal when a manuscript is submitted. 
\author[1]{Ozan Baris Mulayim\textsuperscript{*}}
\author[2]{Pengrui Quan\textsuperscript{*}}
\author[2]{Liying Han}
\author[3]{Xiaomin Ouyang}
\author[4]{Dezhi Hong}
\author[1]{Mario Bergés}
\author[2]{Mani Srivastava}

\authormark{Mulayim and Quan \textit{et al}.}

\address[*]{The authors contributed equally to this work; their names are listed in alphabetical order by last name.}

\address[1]{\orgdiv{Department of Civil and Environmental Engineering}, 
  \orgname{Carnegie Mellon University}, 
  \orgaddress{\city{Pittsburgh}, \postcode{15217}, \state{PA}, \country{USA}}. \email{omulayim@andrew.cmu.edu, marioberges@cmu.edu}}

\address[2]{\orgdiv{Department of Electrical and Computer Engineering}, 
  \orgname{University of California, Los Angeles}, 
  \orgaddress{\city{Los Angeles}, \postcode{90025}, \state{CA}, \country{USA}}. \email{prquan@g.ucla.edu, liying98@ucla.edu, mbs@ucla.edu}}

\address[3]{\orgdiv{Department of Computer Science and Engineering}, 
  \orgname{Hong Kong University of Science and Technology}, 
  \orgaddress{\city{Hong Kong}, \country{Hong Kong}}. \email{xmouyang@cse.ust.hk}}

\address[4]{\orgname{Amazon}, 
  \orgaddress{\city{Seattle}, \state{WA}, \country{USA}}. \email{hondezhi@amazon.com}}

\authormark{Mulayim and Quan \textit{et al}.}



\abstract{

Building energy management (BEM) tasks require processing and learning from a variety of time-series data. Existing solutions rely on bespoke task- and data-specific models to perform these tasks, limiting their broader applicability. Inspired by the transformative success of Large Language Models (LLMs), Time-Series Foundation Models (TSFMs), trained on diverse datasets, have the potential to change this. Were TSFMs to achieve a level of generalizability across tasks and contexts akin to LLMs, they could fundamentally address the scalability challenges pervasive in BEM. To understand where they stand today, we evaluate TSFMs across four dimensions: (1) generalizability in zero-shot univariate forecasting, (2) forecasting with covariates for thermal behavior modeling, (3) zero-shot representation learning for classification tasks, and (4) robustness to performance metrics and varying operational conditions. Our results reveal that TSFMs exhibit \emph{limited} generalizability, performing only marginally better than statistical models on unseen datasets and modalities for univariate forecasting. Similarly, inclusion of covariates in TSFMs does not yield performance improvements, and their performance remains inferior to conventional models that utilize covariates. While TSFMs generate effective zero-shot representations for downstream classification tasks, they may remain inferior to statistical models in forecasting when statistical models perform test-time fitting. Moreover, TSFMs forecasting performance is sensitive to evaluation metrics, and they struggle in more complex building environments compared to statistical models. These findings underscore the need for targeted advancements in TSFM design, particularly their handling of covariates and incorporating context and temporal dynamics into prediction mechanisms, to develop more adaptable and scalable solutions for BEM.

}

\end{Frontmatter}

\section*{Impact Statement}

The insights gained from this study extend beyond building energy management, offering a broader evaluation of Time-Series Foundation Models (TSFMs) when applied to domains they were not explicitly trained on. Our findings serve as a critical indicator of TSFMs’ strengths and limitations in out-of-domain generalization. Their ability to extract zero-shot representations proves remarkably effective, highlighting their potential for applications requiring automated feature learning and transferability across diverse time series tasks. However, their reliance on univariate forecasting and the inability to leverage covariates effectively underscore fundamental architectural limitations. The inability of current strategies to integrate external factors signals the need for new model designs that can better capture contextual dependencies. By demonstrating these gaps, this work informs future developments in TSFMs, guiding research toward more adaptable, scalable, and context-aware models for both energy management and broader time-series applications.

% Some math journals (FLO) require a table of contents. Comment out this line if no ToC is needed.
\localtableofcontents

\section{Introduction}

Building energy management often requires accurate results from a variety of computational tasks performed on time-series data such as forecasting, classification, data imputation, and others. Traditionally, these tasks are addressed using a variety of approaches, including statistical models \citep{zhao2012review}, machine learning \citep{seyedzadeh2018machine} and physics-based techniques \citep{chen2022physical, sun2020review}. However, heterogeneity—in terms of structure, operation, and environmental conditions—between buildings necessitates the development of task-, modality\footnote{In this work, "modality" refers to distinct types of time-series data, whether representing the same phenomenon measured at different time scales or from different vantage points (e.g., indoor vs. outdoor temperature) or entirely different phenomena (e.g., electricity consumption vs. temperature).}-, and building-specific models \citep{blum2019practical}. This fragmented approach introduces challenges in generalizability, limiting the broader applicability of the resulting solutions.

Time-Series Foundation Models (TSFMs) have recently emerged as a promising approach for achieving generalizability across datasets and tasks, drawing inspiration from the transformative success of Large Language Models~(LLMs). Most TSFMs adopt the well-known transformer architecture, incorporating modifications to better accommodate the characteristics of time-series data. These models are typically pretrained on a diverse corpus of time-series datasets, enabling them to perform tasks such as forecasting and anomaly detection in a zero-shot manner. Pioneering examples, such as \texttt{MOMENT} \citep{goswami_moment_2024} and \texttt{TimesFM} \citep{das_decoder-only_2024}, aim to demonstrate generalization across multiple datasets and modalities. Unlike the conventional definitions of foundation models \citep{bommasani2021opportunities}, the vast majority of the TFSMs proposed so far are not designed to generalize to previously unseen tasks. They are, instead, designed to generalize to new data sources of previously seen modalities, and, in some cases, to new modalities that can be represented as time series—while remaining constrained to a fixed set of tasks \citep{baris2025foundationmodelscpsiotopportunities}. 

Though TSFMs may not yet offer the level of generalizability expected from \textit{true} foundation models, we have yet to empirically and systematically evaluate their ability to perform beyond the limits of what they were designed for and trained on. In this work, \emph{we aim to provide a multifaceted zero-shot evaluation of pretrained TSFMs for various building energy management tasks.} Firstly, we analyze the \emph{generalizability} across datasets and modalities for zero-shot univariate forecasting in two key contexts for predictive building management: electricity usage and indoor air temperature. This initial assessment by the authors served as inpsiration for the present manuscript, after results were presented at a conference~\citep{mulayim2024time}. The choice of univariate forecasting for the initial analysis was due to the fact that most TSFMs are only designed to perform univariate forecasting. As a first extension to that early work, in this paper we examine the \emph{covariate utilization} of TSFMs, given the dependence of building energy management tasks on external factors. Specifically, we investigate modeling of building thermal behaviors by comparing the forecasting performance of TSFMs with covariates against univariate predictors and conventional thermal modeling approaches. These comparisons are conducted across various prediction horizons to assess their efficacy under different forecasting requirements. Thirdly, we investigate their performance on extracting meaningful \emph{representations} in a zero-shot setting for classification tasks. Our analysis centers on two distinct classification challenges: appliance classification and metadata mapping. Lastly, we conduct a robustness analysis to understand the details of how TSFMs make decisions. We specifically analyze model's: (1) \emph{stability} under different performance metrics, and (2) \emph{adaptability} to various operational conditions. For this analysis, we again focus on univariate forecasting, as it remains the primary task supported by existing TSFMs.

Our evaluation of TSFMs reveals the following five key findings: First, in terms of \emph{generalizability}, TSFMs exhibit strong performance on datasets they were exposed to during pretraining but struggle with unseen data, particularly when compared to statistical models. While some TSFMs can generalize to new electricity datasets with marginal improvement over statistical methods, their performance on unfamiliar modalities, such as indoor air temperature, is inconsistent—surpassing statistical models for longer forecasting horizons but underperforming for shorter ones. Second,~\emph{covariate utilization} does not enhance TSFM performance, likely due to their training being predominantly univariate; in contrast, traditional methods, optimized for autoregressive forecasting, maintain an advantage regardless of covariate inclusion. Third, ~regarding \emph{representations}, \texttt{Chronos} and \texttt{MOMENT} demonstrate the ability to generate meaningful embeddings without dataset-specific training, achieving performance superior to deep-learning-based classifiers but remaining inferior to non-parametric optimization methods, highlighting the considerable potential of pretrained TSFMs in zero-shot classification. Fourth, \emph{stability} analysis reveals that TSFM performance is highly metric-dependent, with models trained to mimic data distributions better preserving temporal patterns and peaks, whereas those optimized for minimizing magnitude-based errors perform better on root mean square error (RMSE) evaluations, emphasizing the importance of aligning model selection with specific task objectives. Finally, our \emph{adaptability} analysis shows that TSFMs exhibit varying strengths depending on building operational conditions—some models perform better under occupied settings with more dynamic patterns, while statistical models excel under unoccupied, cyclic conditions where energy consumption follows repetitive trends. These findings underscore both the promise and limitations of TSFMs for building energy management, highlighting areas where further advancements are needed to improve their robustness and applicability.

The remainder of this paper is structured as follows: Section \ref{sec:background} provides an overview of existing TSFMs, laying the foundation for our analysis. Sections \ref{sec:univariate} through \ref{sec:behavior} follow a unified structure, each presenting datasets, models, experimental setup, metrics, and results to systematically evaluate TSFMs across different tasks. Specifically, Section \ref{sec:univariate} investigates univariate forecasting, assessing TSFMs in a conventional predictive setting. Section \ref{sec:covariates} extends this examination to forecasting with covariates, evaluating whether incorporating additional contextual information enhances performance. Moving beyond forecasting, Section \ref{sec:classification} explores classification tasks, analyzing TSFMs' ability to generate meaningful representations without explicit task-specific training. Section \ref{sec:behavior} shifts focus to robustness analysis, examining how TSFMs capture temporal dynamics and identifying conditions where they underperform. Section \ref{sec:discussion} synthesizes these findings, discussing their broader implications, and finally, Section \ref{sec:conclusions} concludes the paper with key insights and suggested future research directions.
\section{Background}
\label{sec:background}

\begin{figure}
    \centering
    \includegraphics[width=0.75\linewidth]{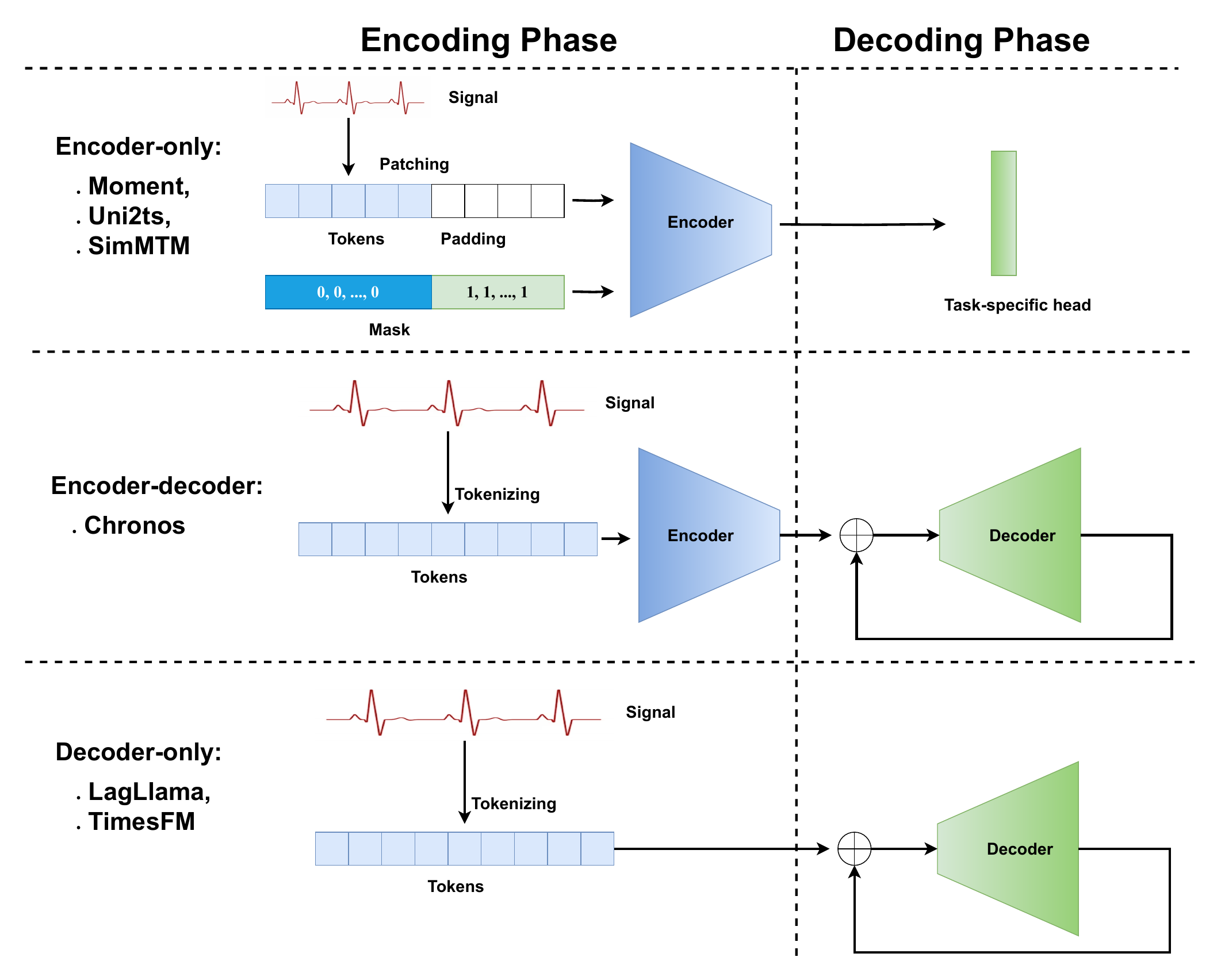}
    \caption{Overview of different TSFM architectures. The task-specific heads enable adaptation to different downstream tasks. The encoder generates latent representations, while the decoder autoregressively predicts future tokens. The encoder and decoder architectures studied in this work are transformer-based but can be generalized to other model architectures}
    \label{fig:architecture}
\end{figure}

\subsection{Existing Time Series Foundation Models}

A foundation model is defined as ``any model that is trained on broad data (generally using self-supervision at scale) that can be adapted (e.g., fine-tuned) to a wide range of downstream tasks'' \citep{bommasani2021opportunities}. Following that definition, we would expect TSFMs to generalize to a wide range of downstream tasks, yet most existing TSFMs either solely focus on forecasting \citep{ansari2024chronos, rasul_lag-llama_2024, das_decoder-only_2024}, require dataset-specific training \citep{jin_time-llm_2024, dong_simmtm_2023}, or can only perform a fixed range of tasks \citep{garza_timegpt-1_2024, goswami_moment_2024}. Given these constraints, understanding the architectural choices and design trade-offs of current TSFMs is essential for evaluating their generalization potential. As this is a nascent and evolving field, with most models released in 2024, we first review existing TSFMs and summarize their architectures and attributes. Figure \ref{fig:architecture} shows the overview of TSFM architecture studied in this work.

\noindent\textbf{Encoder-based architecture.} \texttt{MOMENT} uses an encoder and lightweight prediction heads as backbone~\citep{goswami_moment_2024}. The model tokenizes input data using fixed-length patches and employs transformers~\citep{vaswani2017attention} for prediction, incorporating reversible instance normalization for re-scaling and centering time-series. This approach allows \texttt{MOMENT} to be adapted to various downstream tasks. \texttt{Chronos}  \citep{ansari2024chronos} uses the T5 architecture \citep{raffel2020exploring} for probabilistic forecasting. \texttt{Chronos}  tokenizes time-series values using scaling and quantization and then trains existing transformer-based language model architectures on these tokenized time-series using cross-entropy loss. \texttt{SimMTM}  \citep{dong_simmtm_2023} uses an encoder-based transformer architecture with modules for masking, representation learning, similarity learning, and reconstruction. \texttt{Uni2TS}  \citep{woo_unified_2024} utilizes encoder-only transformers for multivariate time-series, handling different patches and variates. It addresses cross-frequency learning, covariate handling, and probabilistic forecasting. \texttt{UniTime}  \citep{liu_unitime_2024} uses an encoder-based transformer, incorporating semantic instructions through a language encoder to handle domain confusion. 

\noindent\textbf{Decoder-only architecture.} \texttt{LagLlama}  \citep{rasul_lag-llama_2024} uses the Llama architecture \citep{touvron2023llama} for multivariate time-series with a focus on probabilistic forecasting. It employs lag features, data augmentation, and the conventional Llama architecture for robust predictions. \texttt{TimesFM}~\citep{das_decoder-only_2024} model employs a decoder-based transformer for multivariate time-series, processing patches through residual blocks to generate tokens for forecasting. 

\noindent\textbf{Others.} \texttt{TimeLLM}  \citep{jin_time-llm_2024} reprograms an embedding-visible language foundation model, such as Llama and GPT-2 models for univariate time-series forecasting by transforming data into a text format suitable for language models. \texttt{TimeGPT}  \citep{garza_timegpt-1_2024}  leverages a transformer-based architecture. \texttt{TimeGPT}  also deals with missing data, irregular timestamps, uncertainty quantification, fine-tuning, and anomaly detection. 

Table \ref{table:attributes} summarizes the attributes of the above models. Zero-shot means whether models can make predictions without any fine-tuning or they are available without the need for training. We observe that most of these models cannot handle covariates or irregular time-series data (i.e., sampling rate varies over time). Though such attributes do impact building operations, due to limited model availability, in this paper we only focus on the six models with zero-shot abilities and making univariate predictions.

\begin{table}[t]
\centering
\caption{Comparison of TSFM Attributes}
\begin{tabular}{lccccccc}
\toprule
    \textbf{\begin{tabular}[c]{@{}l@{}}Models\end{tabular}} & \textbf{\begin{tabular}[c]{@{}l@{}}Zero \\ Shot\end{tabular}} & \textbf{\begin{tabular}[c]{@{}l@{}}Multi \\ Resolution\end{tabular}}  &  \textbf{\begin{tabular}[c]{@{}l@{}}Covariate \\ Handling\end{tabular}}  & \textbf{\begin{tabular}[c]{@{}l@{}}Irregular \\ Time-series\end{tabular}} & \textbf{\begin{tabular}[c]{@{}l@{}}Multi \\ Task\end{tabular}} & \textbf{\begin{tabular}[c]{@{}l@{}}Takes \\ Timestamps\end{tabular}} \\
\midrule
\texttt{TimeLLM} (2024) & $\xm$ & $\cm$ & $\cm$ & $\xm$ & $\cm$ & $\xm$ \\

\texttt{Uni2TS} (2024) & $\cm$ & $\cm$ & $\cm$ & $\xm$ & $\xm$ & $\cm$ \\

\texttt{SimMTM} (2023) & $\xm$ & $\cm$ & $\xm$ & $\xm$ & $\xm$ & $\xm$ \\

\texttt{TimeGPT} (2024) & $\cm$ & $\cm$ & $\cm$ & $\cm$ & \cm & $\cm$ \\

\texttt{Chronos} (2024) & $\cm$ & $\cm$ & $\xm$ & $\xm$ & $\xm$ & $\xm$ \\

\texttt{MOMENT} (2024) & $\cm$ & $\cm$ & $\xm$ & $\xm$ & $\cm$ & $\xm$ \\

\texttt{LagLlama} (2024) & $\cm$ & $\cm$ & $\xm$ & $\xm$ & $\xm$ & $\cm$ \\

\texttt{TimesFM} (2024) & $\cm$ & $\cm$ & $\cm$ & $\xm$ & $\xm$ & $\xm$ \\

\texttt{UniTime} (2024) & $\xm$ & $\cm$ & $\xm$ & $\xm$ & $\xm$ & $\xm$ \\
\bottomrule
\end{tabular}
\label{table:attributes}
\end{table}

\subsection{Investigating TSFMs in the building domain}  

\noindent\textbf{Leveraging Language Models for Energy Load Forecasting.} Although they did not utilize TSFMs, \citet{xue2023utilizing} introduced a novel approach to energy load forecasting that leverages language models by transforming numerical energy consumption data into natural language descriptions through prompting techniques. Rather than fine-tuning, their method relied on pre-trained transformer-based models—BART, BigBird, and PEGASUS—combined with an autoregressive generation mechanism to iteratively predict future energy loads across multiple horizons. Experimental results on real-world datasets demonstrated that language models, when prompted effectively, could outperform traditional numerical forecasting methods in several cases, highlighting their potential to capture complex temporal dependencies and generalize to new buildings without additional training.  

\noindent\textbf{Improving TSFM Performance in Building Energy Forecasting.} \citet{parkprobabilistic} investigated the use of TSFMs for probabilistic forecasting in building energy systems, evaluating models such as Uni2TS, TimesFM, and Chronos on real-world, multi-month building energy data, including HVAC energy consumption, occupancy, and carbon emissions. Their findings indicated that while zero-shot TSFM performance lagged behind task-specific deep learning models, fine-tuning substantially improved accuracy, in some cases reducing forecasting errors by over 50\%. These results suggest that fine-tuned TSFMs can outperform conventional deep forecasting models, particularly in their ability to generalize to unseen tasks.  

Similarly, \citet{liang2024enabling} examined the adaptation of similar TSFMs for building energy forecasting, identifying the limitations of straightforward fine-tuning due to the lack of universal energy patterns across buildings. To address this, they introduced a contrastive curriculum learning method that optimizes the ordering of training samples based on difficulty to improve adaptation. Their experiments demonstrated that this approach enhanced zero-shot and few-shot forecasting performance by an average of 14.6\% compared to direct fine-tuning, highlighting the potential of structured adaptation strategies for TSFMs in energy forecasting.  

To the best of our knowledge, these are the only existing studies in this domain. Our work differentiates itself by focusing on multiple time-series tasks, including univariate forecasting, forecasting with covariates, and classification, without evaluating fine-tuning performance. Instead, we aim to assess the zero-shot capabilities of TSFMs to determine whether they can offer comparable generalization to existing methods. Additionally, we incorporate building-specific baselines alongside classical statistical models. The goal of this study is not to propose a TSFM-based approach that surpasses existing tools but rather to critically analyze whether TSFMs, as they exist today, can provide sufficient zero-shot generalizability to replicate established forecasting methods.
\section{Univariate Forecasting}
\label{sec:univariate}
As most existing TSFMs were mainly trained to perform univariate forecasting, we pay special attention to understand their performance in this setting. This analysis aims to provide an initial understanding of how these models perform over longer horizons encompassing seasonal variations and diverse household behaviors, specifically focusing on temperature and electricity consumption.

We should note that temperature and electricity data can have distinct time constants based on the source they are collected from. Our initial expectation is that FMs would perform better for electricity since they have observed a similar phenomenon (i.e. electricity measurements from single households) during training. Though they have also observed the temperature from datasets such as Weather and Electricity Transformer Temperature \citep{haoyietal-informer-2021}, the phenomena captured in these datasets exhibit different dynamics than indoor air temperature.

Table \ref{table:familiarity} summarizes the data familiarity and model structures for the studied TSFMs. We use three categories for data familiarity: (1) familiar with the data \textit{modality}: the model is trained with data from the same modality as the test set; (2) familiar with the \textit{dynamics}: the model's training corpus included time-series data generated by dynamical processes similar to that governing the test data; and (3) familiar with the \textit{dataset}: occurring when the model has been trained on the same dataset, which in our case happens for the UCI electricity dataset \citep{misc_electricityloaddiagrams20112014_321} in some models. Table ~\ref{table:familiarity} also captures differences in the model architectures and objective functions. It is worth noting that since \texttt{TimeGPT} is a commercial model, its training data is not publicly available.

\begin{table}[t]
\centering
\caption{Data Familiarity and Model Structures}
\begin{tabular}{p{0.1\textwidth}p{0.25\textwidth}p{0.25\textwidth}p{0.05\textwidth}p{0.15\textwidth}}
\toprule
\textbf{Models} & Level of familiarity with electricity &  Level of familiarity with temperature & Obj. & Transformer architecture \\
\midrule
\texttt{Uni2TS}  & Modality, Dynamics & Modality  & NLL & Encoder \\
\texttt{Chronos}  & Modality, Dynamics, Dataset & Modality & CE & Encoder-decoder \\
\texttt{MOMENT} & Modality, Dynamics, Dataset & Modality & MSE & Encoder \\
\texttt{LagLlama}  & Modality, Dynamics, Dataset & None & NLL & Decoder\\
\texttt{TimesFM}  & Dataset & Modality & MSE & Decoder \\
\texttt{TimeGPT}  & ? & ? &? & ? \\
\bottomrule
\multicolumn{5}{l}{\footnotesize MSE: Mean Squared Error, NLL: Negative Log Likelihood, CE: Cross-Entropy}
\end{tabular}
\label{table:familiarity}
\end{table}

\subsection{Datasets}

We use three main datasets for our experiments:

\noindent \textbf{\textit{ecobee} DYD Dataset}: To test the general ability of TSFMs in predicting indoor temperature, we utilized a large publicly available dataset from \textit{ecobee} \citep{osti_1854924}. This dataset comprises data from 1,000 houses across four U.S. states: California, Texas, Illinois, and New York, sampled at 5-minute intervals with a 1°F resolution during 2017. To ensure statistically significant, yet computationally feasible tests, we selected eight houses with the least number of missing thermostat temperature values from each state, resulting in total of 32 houses. A starting point was randomly sampled from each month, using the same starting points across models for a deterministic comparison, while resampling starting points for each house to ensure greater time diversity. This approach allows us to capture diverse house behaviors, climates, and seasonal variations. Sampling starting points were mainly necessary because the data duration changes based on varying context windows and prediction horizon values.

\noindent \textbf{UCI Electricity Data}: Similarly, to evaluate the general capability of these models in energy consumption predictions, we used the UCI Electricity Load Diagrams dataset \citep{misc_electricityloaddiagrams20112014_321}. This dataset, frequently used in previous works for evaluation, provides an opportunity to reconsider the performance rankings of TSFMs using different metrics. The dataset records electricity consumption in Watts for 370 Portuguese clients from 2011 to 2014, sampled at 15-minute intervals. We randomly sampled 30 houses in this dataset, and for each season, we sampled a starting point, resulting in 16 starting points for each client. This method ensures a comprehensive evaluation across different seasonal contexts and house specifics.

\noindent \textbf{Smart*}: This dataset contains whole-house electricity consumption for 114 single-family apartments for the period of 2014-2016 in kW \citep{barker2012smart}, sampled every 15 minutes. Similar to the previous approach, we randomly selected 30 houses in this dataset and, for each season, sampled 4 different starting points.
% UCI, ecobee, UMass

\subsection{Models}
\label{sec:univariate_models}
\textbf{TSFMs.} We test TSFMs that can do zero-shot univariate forecasting, which leaves us with \texttt{MOMENT}, \texttt{Chronos}, \texttt{TimesFM}, \texttt{TimeGPT}, and \texttt{Lagllama}.

\textbf{Baselines.} Consistent with previous studies, for baselines, we focus on statistical models. However, contrary to the conventions of calibrating statistical models on the available data before testing them, we perform test-time fitting. In other words, the model parameters are obtained by calibrating the model on just the samples contained in the context window of $C$ number of samples. We use this test-time fitting approach for both univariate and covariate forecasting settings. Our baseline models are: \begin{itemize}
    \item \textbf{\texttt{AutoARIMA}}, a variant of ARIMA that automates the selection of the optimal parameters for time-series forecasting by evaluating various combinations \citep{pmdarima}, simplifying the configuration process. The parameters (p, d, q), representing the lag order, the degree of differencing, and the moving average window, respectively, are automatically determined by \texttt{AutoARIMA}, attempting an optimal fit to the data.
    \item \textbf{Seasonal ARIMA (\texttt{S-ARIMA})} extends ARIMA to handle seasonal data patterns, incorporating additional seasonal components: Seasonal Auto-Regressive (SAR), Seasonal Integrated (SI), and Seasonal Moving Average (SMA) \citep{pmdarima}. SARIMA is defined by (p, d, q) for non-seasonal components and (P, D, Q, m) for seasonal components, where m is the number of periods per season.
    \item \textbf{Best Fit Curve (\texttt{BestFit})}, developed to model data trends through curve fitting, offers flexibility by allowing various functional forms tailored to specific dataset characteristics. In our case, we fit a 5\textsuperscript{th} polynomial function: \( f(t) = at^5 + bt^4 + ct^3 + dt^2 + et + f \), where \( a \), \( b \), \( c \), \( d \), \( e \), and \( f \) are coefficients determined during the fitting process and \( t \) represents time.
\end{itemize}

\subsection{Experimental Setting}
\label{sec:experiment_design}
We define the following notations employed throughout this paper:
\begin{itemize}
    \item $H$: Prediction Length (number of samples)
    \item $C$: Context Length (number of samples)
    \item $f_s$: Sampling Interval (minutes)
    \item $D$: Context Duration (hours), defined as $D = \frac{C \cdot f_s}{60}$
    \item $P$: Prediction Duration, also called  \textit{Horizon} (hours), defined as $P = \frac{H \cdot f_s}{60}$
\end{itemize}

While previous literature typically presents results in terms of the number of prediction steps/samples, we express these intervals in terms of hours since it is (a) more intuitive for electricity and temperature predictions, and (b) more comprehensible, particularly as we resample data to analyze performance across various durations.

We selected each context-prediction duration pair $(D, P)$ based on two criteria: $C < 512$, and $H < 64$. The primary rationale behind this choice is that most models are optimized to make predictions within these limits \citep{goswami_moment_2024, ansari2024chronos, das_decoder-only_2024}. During this selection, we considered the sampling rate and resampled to a lower temporal resolution when necessary. The context-prediction-sampling rate tuples $(D\text{(h)}, P\text{(h)}, f_s\text{(mins)})$ are as follows:
\[
\begin{aligned}
    & \text{ecobee:} & & \{(24, 4, 5), (36, 4, 5), (36, 6, 10), \\
    & & & (48, 6, 10), (48, 12, 15), (96, 12, 15), (168, 24, 30)\} \\
    & \text{Smart* and UCI:} & & \{(48, 12, 15), (72, 12, 15), (96, 12, 15), \\
    & & & (48, 24, 15), (72, 24, 15), (96, 24, 30), (168, 24, 30)\}
\end{aligned}
\]

For the Smart* and UCI datasets, we started with a larger number of horizons due to the original sampling rate of UCI being 15 minutes. We maintained this rate for the Smart* dataset to ensure a fair comparison between the two datasets. Furthermore, considering common weekly patterns in household behavior, our aim was to test the prediction performance of these models when a week of data is provided to predict the next day. Hence, we introduced an additional tuple (i.e. $(168, 24, 30)$) to account for this pattern.

\subsection{Metrics}
We used RMSE to evaluate the performance of each model, in line with previous work in this area \citep{goswami_moment_2024, ansari2024chronos}.

\subsection{Results}

\begin{table*}
\small
\caption{RMSE Values for General Analysis. The best is \textbf{bold}, and the second best is \underline{underscored}}
\label{table:consolidated-rmse}
\resizebox{0.90\textwidth}{!}{%
\begin{tabular}{lrrrrrrrrrrr}  % Adjusted the number of columns to match the header removal
\toprule
\textbf{Data} & \multicolumn{2}{c}{\textbf{Parameters}} & \multicolumn{9}{c}{\textbf{Models}} \\
\cmidrule(lr){2-3} \cmidrule(lr){4-12}
& \textbf{$D$} & \textbf{$P$} & \textbf{\texttt{AutoARIMA}} & \textbf{\texttt{S-ARIMA}} & \textbf{\texttt{BestFit}} & \textbf{\texttt{MOMENT}\textsuperscript{*}} & \textbf{\texttt{Chronos}\textsuperscript{*}} & \textbf{\texttt{TimeGPT}} & \textbf{\texttt{LagLlama}\textsuperscript{*}} & \textbf{\texttt{TimesFM}\textsuperscript{*}} & \textbf{\texttt{Uni2TS}} \\
\midrule
\multirow{7}{*}{\rotatebox{90}{(a) UCI (Watts)}} 
& 48 & 12 & 276.6 & 409.4 & 140.0 & 127.6 & \textbf{78.54} & 192.6 & 279.4 & \underline{105.2} & 204.4 \\
& 48 & 24 & 347.3 & 638.5 & 138.6 & 178.2 & \textbf{93.83} & 204.4 & 376.8 & \underline{120.7} & 275.8 \\
& 72 & 12 & 241.1 & 375.4 & 153.6 & 143.3 & \textbf{68.96} & 190.4 & 273.3 & \underline{87.88} & 147.8 \\
& 72 & 24 & 269.0 & 562.3 & 172.7 & 179.4 & \textbf{80.31} & 200.7 & 359.4 & \underline{95.45} & 187.8 \\
& 96 & 12 & 221.8 & 271.7 & 175.4 & 149.8 & \textbf{69.67} & 193.2 & 220.0 & \underline{89.17} & 123.5 \\
& 96 & 24 & 210.1 & 209.4 & 187.3 & 205.9 & 81.18 & 185.6 & 338.5 & \underline{80.87} & \textbf{79.07} \\
& 168 & 24 & 151.4 & 148.1 & 190.7 & 218.4 & \textbf{71.71} & 182.4 & 208.4 & 75.42 & \underline{72.24} \\
\midrule
\multirow{7}{*}{\rotatebox{90}{(b) Smart* (Watts)}} 
& 48 & 12 & 913 & 1,053 & 965 & \underline{907} & 1,076 & 929 & 1,059 & \textbf{897} & 145,228 \\
& 48 & 24 & 936 & 1,097 & 970 & \underline{932} & 1,113 & 956 & 1,148 & \textbf{925} & 1,699,758 \\
& 72 & 12 & \underline{893} & 1,019 & 956 & 899 & 1,030 & 925 & 993 & \textbf{886} & 339,396 \\
& 72 & 24 & \underline{921} & 1,034 & 972 & 926 & 1,071 & 951 & 1,092 & \textbf{914} & 305,820 \\
& 96 & 12 & \underline{874} & 1,022 & 952 & 884 & 1,029 & 909 & 948 & \textbf{869} & 18,741 \\
& 96 & 24 & 775 & 842 & 840 & 787 & 879 & \underline{765} & 944 & \textbf{763} & 734,520 \\
& 168 & 24 & \underline{746} & 830 & 832 & 765 & 830 & 748 & 829 & \textbf{735} & 340,623 \\
\midrule
\multirow{7}{*}{\rotatebox{90}{(c) \textit{ecobee} (°F)}} 
& 24 & 4 & \textbf{1.047} & 1.194 & 1.448 & \underline{1.069} & 1.532 & 1.145 & 1.968 & 1.154 & 1.451 \\
& 36 & 4 & \textbf{1.124} & \underline{1.129} & 2.635 & 1.252 & 1.679 & 1.217 & 2.017 & 1.213 & 1.289 \\
& 36 & 6 & 1.376 & 1.582 & 2.686 & \underline{1.276} & 1.891 & \textbf{1.246} & 2.181 & 1.390 & 1.522 \\
& 48 & 6 & 1.240 & 1.465 & 1.973 & \underline{1.193} & 1.709 & \textbf{1.140} & 2.033 & 1.204 & 1.400 \\
& 48 & 12 & 1.826 & 2.556 & 1.983 & \underline{1.432} & 1.911 & 1.568 & 2.243 & \textbf{1.319} & 1.897 \\
& 96 & 12 & 1.824 & 2.243 & 2.154 & \underline{1.512} & 1.732 & 1.625 & 2.033 & \textbf{1.236} & 1.657 \\
& 168 & 24 & 1.685 & 1.673 & 2.337 & 1.830 & 1.870 & 1.719 & 2.343 & \textbf{1.379} & \underline{1.637} \\
\bottomrule
\multicolumn{12}{l}{\footnotesize \textsuperscript{*}Models that have seen the UCI dataset in their training phase} \\
\end{tabular} 
}
\end{table*}

\begin{figure}[t]
    \centering
        \includegraphics[width=0.8\textwidth]{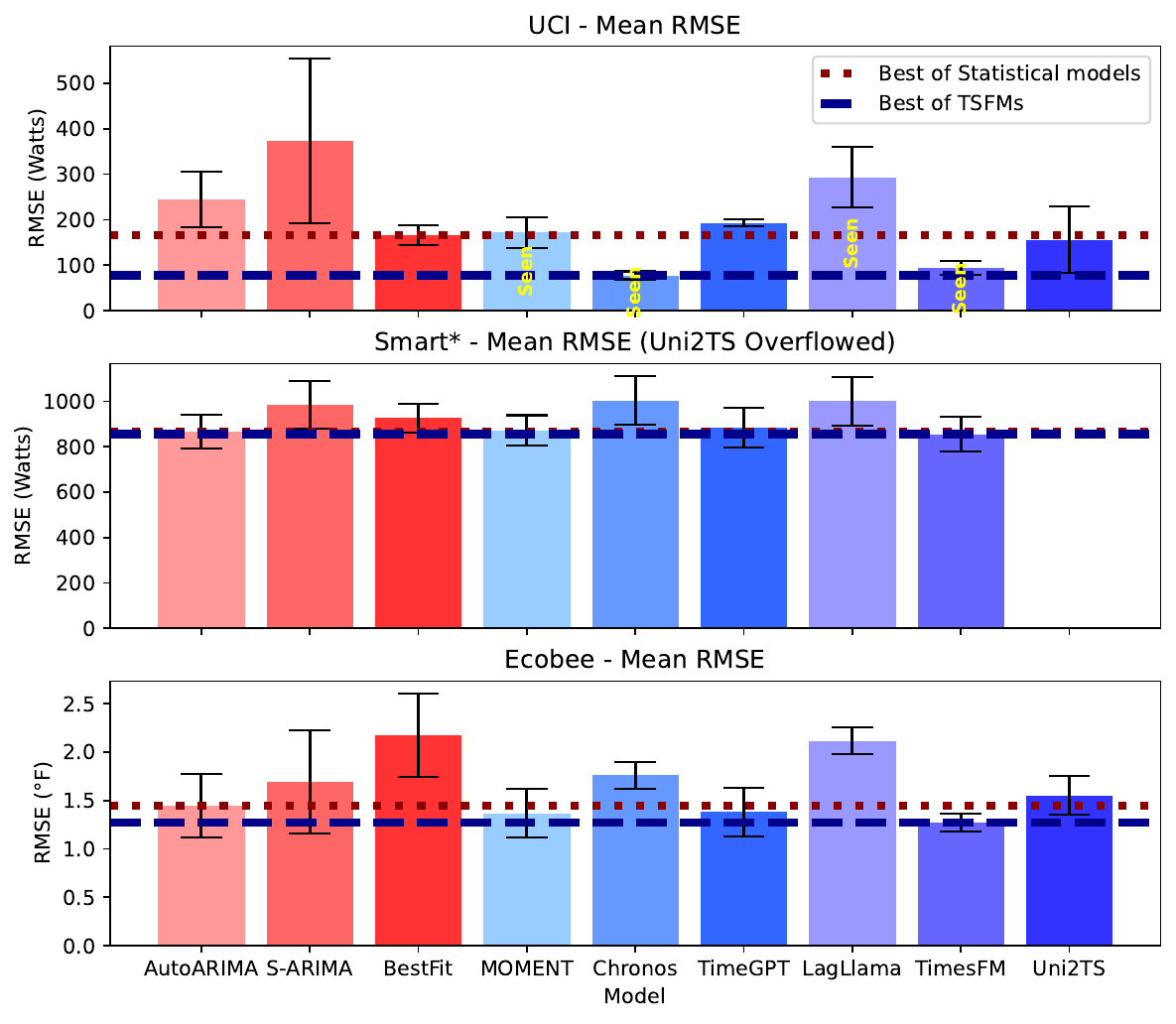} \\
    \caption{Distribution of RMSE values for the three datasets, averaged across varying duration-horizon pairs}
    \label{fig:univariate_results}
\end{figure}

With the goal of understanding the long-term general performance of TSFMs in predicting electricity usage and indoor temperature, we conducted an analysis using three large datasets. Table \ref{table:consolidated-rmse} presents the results for each model, measured by RMSE, while Figure \ref{fig:univariate_results} visualizes these results by illustrating the distribution of RMSE values across all duration-horizon pairs. Our findings offer insights into model performance across a diverse range of seasons and household characteristics.

Our first dataset, UCI as shown in Table \ref{table:consolidated-rmse} (a), has been incorporated into the training process for several of the TSFMs that we evaluated. Comparing the performance of these models on a familiar dataset to their performance on an unfamiliar dataset provides insight into their ability to generalize based on the learned or memorized dynamics. Three of the four models that have previously seen this dataset during training (\texttt{Chronos}, \texttt{TimesFM} and \texttt{MOMENT}) are the top three performing ones particularly for shorter durations. An interesting observation arises with the \texttt{Uni2TS} model, which approaches the performance of the others as the duration increases, though it has not seen this data before. This phenomenon may occur because the data are resampled as the duration increases, resulting in a dataset that differs slightly from the original training data sampled every 15 minutes. Consequently, \texttt{Uni2TS} manages to close the performance gap. Comparing statistical models with foundation models, we observed that BestFit outperforms models that have not been trained on the UCI dataset (\texttt{TimeGPT} and \texttt{Uni2TS}) on $C = 48$ and $72$. Besides, \texttt{AutoARIMA} and \texttt{BestFit} outperform \texttt{LagLlama} consistently despite \texttt{LagLlama} being trained on the dataset.

Upon switching to the Smart* electricity dataset (shown in Table \ref{table:consolidated-rmse} (b)), we observe a shift in comparative model performance. It is important to note that the number of steps is not strictly different for shorter and longer durations as we resample the data for longer durations. Thus, performance discrepancies can be attributed solely to the behavioral characteristics during those durations. Evaluated on the Smart* dataset, \texttt{Chronos} loses its leading position to \texttt{TimesFM}. \texttt{MOMENT} and \texttt{TimeGPT} are comparable in terms of absolute performance, following \texttt{TimesFM} and \texttt{AutoARIMA}. Other FMs are consistently worse than the previous three TSFMs. Specifically, Uni2TS produces forecasting outliers, leading to exceptionally large errors. Regarding statistical models, overall, \texttt{AutoARIMA} performs consistently close to the best TSFM \texttt{TimesFM}, \texttt{BestFit} outperforms three foundation models, and \texttt{S-ARIMA} outperforms two foundation models. The considerably larger errors, compared to those observed with the UCI dataset, are likely due to a combination of two factors: (1) the intrinsic predictability of Smart* is more challenging, as evidenced by the predictions of statistical models (from ~200 Watts to ~900 watts), and (2) the limited generalizability of TSFMs, as manifested by the diminishing performance gap between statistical models and TSFMs compared to the UCI dataset. This indicates a limitation in the performance of TSFMs without familiarity with the data set. 

The \textit{ecobee} dataset presents a more diverse set of results (Table \ref{table:consolidated-rmse}(c)), with statistical models outperforming others for shorter durations. As the duration increases, temperature variations tend to smooth out, where TSFMs seem to perform better. Nonetheless, \texttt{AutoARIMA} remains competitive, nearly matching the best-performing forecasting models. When comparing \texttt{MOMENT}, \texttt{TimeGPT}, and \texttt{TimesFM}, we find that \texttt{TimesFM} excels in long-duration predictions, whereas \texttt{MOMENT} and \texttt{TimeGPT} perform best during moderate durations where temperature changes are less variable. This nuanced performance indicates the importance of considering duration-specific characteristics when evaluating model efficacy. 

\subsection{Summary}
In summary, as shown in Figure \ref{fig:univariate_results}, the TSFMs show different levels of performance based on prior exposure to the datasets. In the realm of electricity datasets, though \texttt{Chronos} stands out as the most effective model for a familiar electricity dataset (i.e., UCI), this advantage diminishes with a new dataset Smart*. In contrast, models like \texttt{TimesFM} and \texttt{MOMENT} demonstrate a robust capability to generalize to novel data, though their performance improvement over \texttt{AutoARIMA} is marginal. Lastly, statistical models and TSFMs perform closely in indoor air temperature forecasting, with a slight performance gap separating the best statistical models from the TSFMs.

\section{Forecasting with Covariates}
\label{sec:covariates}

Our previous analysis focused on univariate forecasting, as most existing models are designed specifically for this task. However, building energy management inherently relies on multiple covariates, including solar irradiation, outdoor air temperature, and occupancy patterns. Given this dependency, in this section, we evaluate the performance of TSFMs in forecasting with covariates, comparing them against established building thermal modeling baselines.

\subsection{Datasets} 
\textbf{ecobee + solar.} We utilize a subset of the ecobee dataset \citep{osti_1854924}, augmented with solar irradiance data derived from the National Solar Irradiation Database \citep{NREL_NSRDB_API}. This subset was constructed by selecting consecutive 28-day training and 4-day testing periods from the cooling seasons of 1,000 houses, as used in previous studies~\citep{mulayim2024beyond, mulayim2024leveraging}. After filtering for houses with complete and consistent data, the final subset includes 841 houses. 

\noindent \textbf{Covariates.} Accurately modeling building thermodynamics requires a clear understanding of the covariate structure, which is grounded in fundamental physical principles. Variables such as heat gains from occupants and equipment significantly affect thermal dynamics but are often challenging to measure outside controlled experimental conditions. Alternatively, data from smart thermostats, including heating/cooling duty cycles, solar irradiance, and outdoor air temperature, can be effectively mapped to the parameters of thermal resistance-capacitance (RC) networks, a widely adopted approach for modeling building thermodynamics.

In this study, we predict the target variable $T_t$ (°F), an indoor temperature controlled by the HVAC system at time $t$. The covariates utilized for each house are defined as follows:

\begin{itemize}
    \item $T_{oat}(t)$: Outdoor air temperature at time $t$ (°F), which drives heat exchange between the building and its environment.
    \item $Q_{sol}(t)$: Global horizontal irradiance at time $t$ ($W/m^2$), representing the solar heat gain impacting the building envelope.
    \item $\mu_{c}(t)$: Cooling duty cycle at time $t$ (unitless, ranging from 0 to 1), indicating the fraction of time the cooling system is active during a given time step.
\end{itemize}

By formulating these covariates, we align with widely adopted practices in large-scale thermal modeling studies~\citep{vallianos2023application, mulayim2024leveraging}. These measurable and practical covariates ensure the applicability of our approach to real-world scenarios, bridging the gap between theoretical modeling and operational deployments.

\subsection{Models}

\textbf{TSFMs.} We evaluate the forecasting performance of three TSFMs: \texttt{Uni2TS}, \texttt{TimesFM}, and \texttt{TimeGPT}, which are the only models in this study that explicitly support covariate-based predictions. While other models may perform multivariate forecasting, it is important to distinguish that, in many cases, multivariate predictions are merely equivalent to performing multiple univariate forecasts independently, without incorporating the covariate structure into the predictions. The manner in which covariates are utilized varies across the three TSFMs we consider.

First, \texttt{Uni2TS} incorporates covariates by flattening multivariate time series into a single sequence, employing variate encodings and a binary attention bias to disambiguate variates, ensure permutation equivariance/invariance, and scale to arbitrary numbers of variates. Second, though \texttt{TimeGPT} is a commercial and closed-source model, we include it as a baseline for using covariates. Third, unlike the other two methods, \texttt{TimesFM} does not directly perform covariate forecasting. Instead, it follows a two-step approach: first fitting a linear model to capture the relationship between the time series and its covariates, and then applying \texttt{TimesFM} to forecast the residuals.

\noindent \textbf{Baselines.} To evaluate the performance of TSFMs in modeling building thermodynamics, we compare them against three existing approaches. Again, we perform test-time fitting to the following models:\begin{itemize}
    \item \texttt{Ridge}: A ridge regression model, which serves as a representative data-driven baseline for thermal prediction tasks as shown by \citep{mulayim2024beyond, huchuk2022evaluation}.
    \item Random Forest (\texttt{RF}): A machine learning model, which has shown superior performance in a relevant comparison study \citep{huchuk2022evaluation}.
    \item \texttt{R2C2}: A physics-based RC network with 2 capacitances and 2 resistances, a well-established approach for modeling thermal behavior by leveraging physical laws \citep{mulayim2024leveraging}. 
\end{itemize}

\subsection{Experimental Setting}
Both baseline models and TSFMs were tested on the same subset of the ecobee dataset as described above. Initial training dataset for baselines consisted of $N=672$ time steps (28 days), while the test set consisted of $H=96$  time steps (4 days). However, given that TSFMs are typically optimized for making predictions where the context window size $C<=512$ and specifically, \texttt{MOMENT} requires $C+H < 512$. Thus, we made an adjustment where we utilized $C=448$ values from the initial training set for training the baselines and for providing context to the TSFMs. For predictions, we utilized the first $H=64$ values from the test set. This configuration ensures alignment with TSFM design principles while enabling a fair comparison across all models since both baselines and TSFMs have equal information for training/context.

The experimental test is designed around a practical application: Model Predictive Control (MPC). At any given time $t$, the controller is assumed to have perfect historical knowledge of key variables, including $T(t-511:t)$ (thermostat temperature), $T_{oat}(t-511:t)$ (outdoor air temperature), $\mu_c(t-511:t)$ (cooling duty cycle), and $Q_{sol}(t-511:t)$ (solar irradiance). For future predictions, weather data such as $T_{oat}(t+1:t+64)$ and $Q_{sol}(t+1:t+64)$ are provided with added Gaussian noise to account for realistic uncertainty in forecasting. The noise addition follows a systematic process to capture temporal variability and cumulative uncertainty. First, a set of standard deviations, \( \sigma_i \), is selected, representing different levels of noise intensity. The noise addition follows a systematic process to capture temporal variability and cumulative uncertainty. A set of standard deviations, \( \sigma_i \), is selected to represent different levels of noise intensity. For each time step \( t \) into the prediction horizon, noise is sampled as:
\[
n_t \sim \mathcal{N}(0, t \cdot \sigma_i^2),
\]

Here, \( n_t \) represents the cumulative forecast error at time \( t \). This reflects the fact that the sum of \( t \) independent Gaussian noise terms, each distributed as \( \mathcal{N}(0, \sigma_i^2) \), yields a Gaussian distribution with variance \( t \cdot \sigma_i^2 \), and therefore a standard deviation of \( \sqrt{t} \cdot \sigma_i \). The sampled noise \( n_t \) is added to the corresponding weather variable, such as \( T_{oat}(t) \) or \( Q_{sol}(t) \), to simulate uncertainty in future predictions.

By varying \( \sigma_i \), the impact of different noise levels on model performance is systematically evaluated, providing insights into the robustness of the controller and the forecasting models under varying degrees of uncertainty. This approach aligns with realistic operational conditions, where future weather predictions are inherently imperfect.

The cooling duty cycle $\mu_c(t+1:t+64)$, which is determined by an optimization solution (either via sampling or direct convex optimization), represents a controllable variable in the MPC framework. In this study, $\mu_c(t+1:t+64)$ is assumed to be perfectly known, consistent with the assumptions employed in prior research on building thermal behavior modeling. This assumption is not without effects. In practice, MPC-based controllers output $\mu_c(0:P_h)$ (where $P_h$ is the lookahead window of the MPC controller) but only use the first action in the sequence and in the next step, regenerate another action set. Thus, in its practical deployment, this resulting action set can change with each step due to the errors in predictions of next states or forecasts of external variables. However, in our evaluation, the output $\mu_c(t+1:t+64)$ is what shall be deployed no matter the errors being observed throughout the deployment. The \textit{true} practical use of these forecasters in MPC settings, thus, can only be evaluated via experimental studies, which is out scope for this work. While our evaluation provides some insight into how the performance would change with respect to covariate prediction error, it does not include the corrective actions that can be taken by the MPC solver at each timestep.
 
To evaluate the efficacy of TSFMs, we frame our investigation around two core questions:
\begin{itemize}
    \item Impact of Covariates: Does the inclusion of additional covariates enhance the forecasting accuracy of TSFMs compared to univariate approaches for modeling building thermal behavior?
    \item Comparison with Traditional Models: Can TSFMs outperform conventional modeling paradigms, such as ridge regression and RC networks, in forecasting building thermal behavior?
\end{itemize}

\subsection{Metrics}

Unlike baselines that are trained to minimize the error of one-step-ahead predictions, TSFMs are designed to predict multiple time steps ($P > 1$) simultaneously. This capability necessitates a redefinition of the evaluation metric for each question separately to accommodate the architectural features of TSFMs.

To address the first research question, we assess the accuracy of TSFMs across the entire $H=64$ prediction horizon for both univariate and covariate-based settings. The RMSE is used as the primary evaluation metric, calculated over the full prediction interval (shown as RMSE\textsubscript{all}) to quantify the difference between predicted and actual values.

For the second question, we compare model predictions across common time intervals relevant to building thermal modeling. Prior studies have evaluated forecasting accuracy at intervals ranging from 1 hour~\citep{mulayim2024beyond, vallianos2023application, huchuk2022evaluation, mulayim2024leveraging}, to 6 hours~\citep{huchuk2022evaluation}, and up to 24 hours~\citep{vallianos2023application, mulayim2024beyond}. Following this approach, we sample predictions from the TSFMs and baseline models at the 1-hour (RMSE\textsubscript{1st}), 6-hour (RMSE\textsubscript{6th}), and 24-hour (RMSE\textsubscript{24th}) marks. The RMSE for these sampled intervals is computed across all 771 houses in the dataset, providing a comprehensive evaluation of model performance.

\subsection{Results}

\begin{table*}[h!]
    \centering
    \caption{Results Across Different Noise Levels where $D=448$ and $P=64$}
    \label{tab:rmse_covariates}
    \resizebox{\textwidth}{!}{%
    \begin{tabular}{l|cccc|ccc|cccccc}
        \toprule
        & & \multicolumn{3}{c|}{\textbf{Baselines with Covariates}} & \multicolumn{3}{c|}{\textbf{Forecasting with Covariates}} & \multicolumn{6}{c}{\textbf{Univariate Forecasting}} \\
        
        \cmidrule(lr){3-5} \cmidrule(lr){6-8} \cmidrule(lr){9-14}
        \textbf{\( \sigma_i \) } & \textbf{Metric} & \textbf{\texttt{R2C2}} & \textbf{\texttt{Ridge}} & \textbf{\texttt{RF}} & \textbf{\texttt{TimesFM}} & \textbf{\texttt{Uni2TS}} & \textbf{\texttt{TimeGPT}} & \textbf{\texttt{TimesFM}} & \textbf{\texttt{Uni2TS}} & \textbf{\texttt{TimeGPT}} & \textbf{\texttt{Chronos}} & \textbf{\texttt{LagLlama}} & \textbf{\texttt{MOMENT}} \\
        \midrule
        0 & RMSE\textsubscript{all} & 2.84 & \textbf{1.33} & \underline{1.42} & 2.13 & 1.58 & 1.62 & 1.60 & 1.59 & 1.68 & 2.08 & 2.13 & 2.06 \\
        & RMSE\textsubscript{1st} & \textbf{0.35} & \underline{0.36} & 0.36 & 1.64 & 0.44 & 0.75 & 0.46 & 0.44 & 0.42 & 0.53 & 1.44 & 0.65 \\
        & RMSE\textsubscript{6th} & 0.92 & \underline{0.76} & \textbf{0.76} & 1.57 & 0.76 & 0.98 & 0.82 & 0.77 & 0.81 & 1.06 & 1.71 & 0.96 \\
        & RMSE\textsubscript{24th} & 1.52 & \textbf{1.03} & 1.14 & 1.72 & 1.14 & 1.23 & 1.16 & 1.15 & \underline{1.10} & 1.58 & 1.45 & 1.70 \\
        \hline
        0.1 & RMSE\textsubscript{all} & 2.84 & \textbf{1.33} & \underline{1.41} & 2.13 & 1.58 & 1.62 & 1.60 & 1.59 & 1.68 & 2.04 & 2.13 & 2.06 \\
        & RMSE\textsubscript{1st} & \textbf{0.35} & 0.36 & \underline{0.36} & 1.64 & 0.44 & 0.75 & 0.46 & 0.43 & 0.42 & 0.53 & 1.42 & 0.65 \\
        & RMSE\textsubscript{6th} & 0.92 & \underline{0.76} & \textbf{0.76} & 1.57 & 0.79 & 0.98 & 0.82 & 0.77 & 0.81 & 1.01 & 1.74 & 0.96 \\
        & RMSE\textsubscript{24th} & 1.52 & \textbf{1.03} & 1.13 & 1.71 & 1.15 & 1.23 & 1.16 & 1.15 & \underline{1.10} & 1.49 & 1.48 & 1.70 \\
        \hline
        0.2 & RMSE\textsubscript{all} & 2.87 & \textbf{1.35} & \underline{1.43} & 2.14 & 1.58 & 1.63 & 1.60 & 1.59 & 1.68 & 2.06 & 2.14 & 2.06 \\
        & RMSE\textsubscript{1st} & \textbf{0.35} & \underline{0.36} & 0.36 & 1.64 & 0.45 & 0.74 & 0.46 & 0.43 & 0.42 & 0.52 & 1.43 & 0.65 \\
        & RMSE\textsubscript{6th} & 0.92 & \underline{0.76} & 0.77 & 1.57 & 0.78 & 1.00 & 0.82 & \textbf{0.76} & 0.81 & 1.09 & 1.75 & 0.96 \\
        & RMSE\textsubscript{24th} & 1.53 & \textbf{1.04} & 1.12 & 1.73 & 1.15 & 1.24 & 1.16 & 1.14 & \underline{1.10} & 1.48 & 1.44 & 1.70 \\
        \hline
        0.5 & RMSE\textsubscript{all} & 3.00 & \textbf{1.47} & \underline{1.49} & 2.23 & 1.58 & 1.70 & 1.60 & 1.59 & 1.68 & 2.06 & 2.12 & 2.06 \\
        
        & RMSE\textsubscript{1st} & \textbf{0.35} & 0.36 & \underline{0.36} & 1.64 & 0.44 & 0.76 & 0.46 & 0.44 & 0.42 & 0.53 & 1.41 & 0.65 \\
        & RMSE\textsubscript{6th} & 0.92 & \underline{0.77} & 0.77 & 1.58 & 0.78 & 1.00 & 0.82 & \textbf{0.75} & 0.81 & 1.03 & 1.70 & 0.96 \\
        & RMSE\textsubscript{24th} & 1.59 & \textbf{1.10} & 1.17 & 1.78 & 1.15 & 1.28 & 1.16 & 1.12 & \underline{1.10} & 1.48 & 1.43 & 1.70 \\
        \hline
        1 & RMSE\textsubscript{all} & 3.42 & 1.77 & 1.64 & 2.38 & \textbf{1.59} & 1.84 & 1.60 & \underline{1.59} & 1.68 & 2.04 & 2.13 & 2.06 \\
        & RMSE\textsubscript{1st} & \textbf{0.35} & 0.36 & \underline{0.36} & 1.63 & 0.44 & 0.78 & 0.46 & 0.44 & 0.42 & 0.52 & 1.46 & 0.65 \\
        & RMSE\textsubscript{6th} & 0.93 & 0.78 & \textbf{0.76} & 1.61 & 0.78 & 1.06 & 0.82 & \underline{0.77} & 0.81 & 1.00 & 1.75 & 0.96 \\
        & RMSE\textsubscript{24th} & 1.84 & 1.33 & 1.25 & 1.91 & 1.15 & 1.49 & 1.16 & \underline{1.14} & \textbf{1.10} & 1.43 & 1.43 & 1.70 \\
        \bottomrule
    \end{tabular}%
    }
\end{table*}

\textbf{Impact of Covariates.} The inclusion of covariates does not enhance the forecasting accuracy of TSFMs compared to their univariate counterparts, as evidenced by RMSE\textsubscript{all} values in Table \ref{tab:rmse_covariates}. For \texttt{Uni2TS}, the addition of covariates results in only marginal improvements relative to its univariate implementation. However, as noise levels increase, \texttt{Uni2TS} performance begins to degrade, indicating that while it incorporates covariate information, its sensitivity to uninformative or noisy covariates undermines its robustness. Similarly, \texttt{TimeGPT} performs slightly better when $\sigma_i \leq 0.2$, but its performance deteriorates beyond that threshold. This suggests that \texttt{TimeGPT} can utilize covariate information effectively under low-noise conditions, yet the introduction of noisy covariates adversely impacts its accuracy, as expected. \texttt{TimesFM}, which employs an ad hoc approach to covariate integration, consistently underperforms, further highlighting that simply including covariates does not guarantee improved forecasting accuracy, especially when the model lacks a principled design for handling multivariate structures.

Notably, even in the absence of noise, the performance of \texttt{TimeGPT} and \texttt{Uni2TS} with covariates shows only slight improvements over univariate forecasters, while \texttt{TimesFM} performs worse. This may stem from the primary training objective of these models being univariate forecasting, underscoring the need for more effective methods to integrate and utilize covariate information in forecasting tasks.

\textbf{Comparison with Traditional Models.} The results indicate that simple traditional models, such as ridge regression and RC networks, consistently outperform TSFMs in forecasting building thermal behavior, even when trained on the same amount of data as the TSFM context window. This superiority is evident in autoregressive multistep predictions under zero noise conditions (as shown by RMSE\textsubscript{6th} and RMSE\textsubscript{24th}), a task for which these baselines were not explicitly designed. Only under increasing noise levels do \texttt{Uni2TS}'s univariate forecasters outperform \texttt{Ridge} and RandomForest for RMSE\textsubscript{6th}, and \texttt{TimeGPT} achieves better results for RMSE\textsubscript{24th}. For one-step-ahead predictions (RMSE\textsubscript{1st}), which are the primary training objective of the baselines, none of the TSFMs come close to their performance. Among the baselines, \texttt{R2C2} consistently delivers the best one-step-ahead predictions, with minimal performance gaps compared to other traditional methods, underscoring the strength of domain-specific models in capturing the physics and statistical structure of building thermal dynamics.

\subsection{Summary}
TSFMs, regardless of whether they incorporate covariates, generally struggle to match the performance of traditional baselines. Only under conditions of increased noise do univariate TSFM forecasters begin to outperform the baselines in multistep predictions. Overall, the ability of traditional models to achieve superior performance even with very limited training data suggests that TSFMs still have significant room for improvement before they can consistently surpass these established methods in this domain.
\section{Classification}
\label{sec:classification}
Having demonstrated TSFM’s forecasting capabilities, we now explore its performance on additional tasks. While imputation and anomaly detection can often be treated as natural extensions of forecasting—inferring missing values or flagging outliers based on deviations from predicted signals—classification requires a fundamentally different approach, which is precisely why we chose it. 

Unlike forecasting, where a context window directly informs future predictions, classification lacks a straightforward mechanism to provide the model with “example” labels for unseen data. Consequently, zero-shot classification is not inherently possible within TSFMs, as there is no built-in way to infer class labels without supervision. To address this, we leverage \texttt{MOMENT}'s ability to provide representations to extract latent time series embeddings without requiring any labeled training data. These embeddings serve as fixed feature representations, which are then used to train a separate classifier head, such as a Support Vector Machine (SVM), enabling classification without directly finetuning the TSFM itself. To evaluate the performance of this approach, we focus on two distinct yet prevalent classification tasks in building energy management: 
\begin{itemize}
    \item Appliance classification – identifying which appliance is active based on time series data, and
    \item Metadata mapping – assigning standardized semantic labels (e.g., sensor types, setpoints, or alarms) to diverse building data points.
\end{itemize}

In the following sections, we describe how TSFM is adapted and evaluated for these classification scenarios.

\subsection{Datasets}

We leverage two datasets for our experiments on classification:

\textbf{WHITED.} The WHITED dataset \citep{kahl2016whited} is a collection of high-frequency energy measurements capturing the start-up transients of 110 appliances, spanning 47 types across six regions worldwide. After grouping multiple appliance brands under their respective types, this dataset has 56 unique appliance classes. Data were recorded at 44.1 kHz using low-cost sound card-based hardware, focusing on a five-second window that captures the transient characteristics essential for tasks like appliance classification and regional grid analysis.

Each sample in the dataset includes voltage and current measurements. Following \citep{kahl2016whited}, which reports the best performance when power values are considered, we computed the instantaneous power by multiplying voltage and current to obtain a single feature. Finally, we split the dataset 80/20 into training and testing subsets, resulting in 1,071 training samples and 268 testing samples. 

\noindent \textbf{BTS.} The Building TimeSeries (BTS) dataset \citep{prabowo2024bts} comprises over 10,000 time series data points spanning three years across three buildings in Australia. It adheres to the Brick schema \citep{balaji2016brick} for standardization and supports time series ontology classification—mapping each time series to one or more ontological classes (e.g., sensors, setpoints, or alarms). This multi-label classification problem is challenging due to extreme class imbalance and domain shifts across different buildings, offering a strong benchmark for assessing interoperability and scalability in building analytics.

This dataset requires a multilabel classifier since it features 94 sub-classes of labels. Each data point is annotated with:
\begin{itemize}
    \item Positive Labels: The true label and its parent classes
    \item Zero Labels: All subclasses of the true label
    \item Negative Labels: All unrelated labels
\end{itemize}

Since there is an ongoing public competition\footnote{\protect\url{https://www.aicrowd.com/challenges/brick-by-brick-2024}} utilizing this dataset, only a subset is publicly available. We used this subset and split it 80/20 into training and testing, resulting in 25,471 samples for training and 6,368 for testing.

One important consideration is the extent to which \texttt{MOMENT} and \texttt{Chronos} have been exposed to data similar to WHITED and BTS during their pretraining. While \texttt{MOMENT} and \texttt{Chronos} have encountered electricity and temperature data across a range of time constants, the characteristics of these datasets differ significantly from those used in this study. WHITED consists of high-frequency (kHz-level) power measurements captured over a few minutes, whereas  were primarily trained on lower-resolution datasets such as the UCI electricity dataset, which records household power usage at 15-minute intervals. Similarly, the BTS dataset spans multiple sensor modalities—including temperature, alarms, damper positions, and airflow values—most of which are distinct from what these models have seen. The closest match is outdoor air temperature, which aligns more closely with the weather datasets included in the pretraining corpus of \texttt{MOMENT} and \texttt{Chronos}, while the remaining modalities represent unseen or significantly different data distributions. As a result, evaluating \texttt{MOMENT} and \texttt{Chronos} on WHITED and BTS not only provides insight into their applicability for building energy management tasks but also serves as a broader test of their ability to generalize to datasets that deviate from the training distribution.

\subsection{Models}

\textbf{TSFMs.} Among the TSFMs evaluated, \texttt{MOMENT} and \texttt{Chronos} were the only models capable of performing classification tasks. We utilized them as a zero-shot representation learner in combination with a dataset-specific classification head, as recommended in \citet{goswami_moment_2024}. Specifically, \texttt{MOMENT} and \texttt{Chronos} were directly applied without retraining to extract latent representations of the time series data. These latent representations, along with their corresponding class labels, were then used to train a \texttt{SVM}. 

\noindent \textbf{Baselines.} Based on the performance on UCR classification archive reported in \citep{goswami_moment_2024}, we specifically chose the best performing models from each category and trained each model on our selected classification datasets independently. These baselines are:

\begin{itemize}
    \item \texttt{TS2Vec} \citep{yue2022ts2vec} (self-supervised representation learning): Serving as strong baselines in \citep{goswami_moment_2024}, \texttt{TS2Vec} employ a two-stage manner to perform classification: first, it performs contrastive learning to obtain latent representation on the targeted dataset in a self-supervised manner. At the second stage, a classification head such as \texttt{SVM} is trained on the embeddings learned by the \texttt{TS2Vec} model using class labels.
    \item \texttt{ResNet} \citep{ismail2019deep} (supervised deep learning): As a model demonstrated strong performance consistently among nine deep learning classifiers in \citep{ismail2019deep}, we use the \texttt{ResNet}-18 model as a baseline. The model is trained end-to-end with the classification loss and class labels as supervision.
    \item \texttt{DTW}/\texttt{SDTW} \citep{cuturi2017soft} (supervised non-parametric optimization models): Soft-Dynamic Time Warping (\texttt{SDTW}) is a differentiable variant of the traditional \texttt{DTW} algorithm, which is designed to replace the hard minimum operation of \texttt{DTW} with a relaxation. It retains the ability to handle temporal shifts between time series with significant speedup. To perform time series classification, we combine the distances measured by \texttt{SDTW} or \texttt{DTW} to assign a label based on the majority class of the closest samples. We select $K=1$ in the experiments.

    To counteract the non-positivity of \texttt{SDTW}, we use the Soft-DTW divergence as the distance metric $D(\cdot , \cdot)$ for the \texttt{KNN} classifier:
    \begin{equation}
        D(s, \hat{s}) = \text{\texttt{SDTW}}(s, \hat{s}) - \frac{1}{2}(\text{\texttt{SDTW}}(s, s) + \text{\texttt{SDTW}}(\hat{s}, \hat{s}))
    \end{equation}
    where $s$ and $\hat{s}$ are two time series.
\end{itemize}

\subsection{Experimental Setting} 

Our training methodology is structured to accommodate different types of models. (1) For fully supervised models, such as \texttt{SDTW} and \texttt{ResNet}, the training process involves providing the training set with class labels, and the models are trained in an end-to-end manner with categorical cross entropy loss. (2) For models trained in two stages, like \texttt{TS2Vec}, we first use contrastive learning to train a representation learner without class labels. Subsequently, a separate classification head is trained using class labels. During inference, the time series data is passed through the frozen representation learner to generate latent embeddings, which are then input into a pre-trained \texttt{SVM} to predict class labels. (3) For zero-shot representation learners, such as \texttt{MOMENT}, we directly train the \texttt{SVM} with class labels and time series embeddings obtained by \texttt{MOMENT}. During inference, the outputs of the \texttt{MOMENT} model are concatenated with the \texttt{SVM} to produce the final predictions.

For the BTS dataset, we employed a multi-output classifier due to its multilabel requirement, as each time series can belong to arbitrary numbers of classes simultaneously. Specifically, we employed the binary relevance strategy to adapt the models \citep{zhang2018binary}. For non-parametric optimization models such as \texttt{SVM} and \texttt{KNN}, we trained $K$ classifiers where $K$ is the number of classes. For deep learning models such as \texttt{ResNet}, we concatenate a \texttt{sigmoid} function to the output logit of the model to generate the probability of classes. To address the input sequence length limitation of \texttt{MOMENT}, we resampled all time series into 512 steps. 

\subsection{Metrics} 

For the WHITED dataset, we use macro precision, recall, and F1 score to ensure equal importance for all appliance types in this class-balanced dataset. While these metrics provide a comprehensive evaluation, accuracy is the primary metric, aligning with its common use in prior work \citep{kahl2016whited}.

For the BTS dataset, we use micro precision, recall, and F1 score to evaluate performance, as they provide a comprehensive assessment across all classes. Due to the significant class imbalance, the micro F1 score is the most reliable indicator \citep{prabowo2024bts}. The reason is that the predominance of the negative class makes the accuracy misleading, as models can achieve high accuracy by predicting the majority class without effectively distinguishing minority classes.

\subsection{Results}

\begin{table*}[h!]
    \centering
    \caption{Results Across Different Metrics and Datasets. IBL means instance-based learning, E2E indicates that the model is trained directly using a classification loss in an end-to-end manner. SS-RL means self-supervised representation learning. PT-RL means pre-trained representations. SVM/NN means an SVM or a Neural Network (NN) classification head is trained on top of a frozen representation. $N$ is the number of the training samples}
    \label{tab:rmse}
    \resizebox{0.95\textwidth}{!}{%
    \begin{tabular}{l|ccccccc|ccc}
        \toprule
        \textbf{Dataset} & \textbf{Metric} & \textbf{\texttt{DTW}} & \textbf{\texttt{SDTW}} & \textbf{\texttt{ResNet}18} & \textbf{\texttt{TS2Vec}} & \textbf{\texttt{MOMENT}} & \textbf{\texttt{Chronos}} & \textbf{\texttt{TS2Vec}} & \textbf{\texttt{MOMENT}} & \textbf{\texttt{Chronos}}\\
        \midrule
        Method & - & IBL & IBL & E2E & US-RL+SVM\textsuperscript{*} & PT-RL+SVM\textsuperscript{*} & PT-RL+SVM\textsuperscript{*} & US-RL+NN & PT-RL+NN & PT-RL+NN \\
        Complexity & - & $O(N^2)$ & $O(N^2)$ & $O(N)$ & $O(N^2)$ & $O(C)$ & $O(C)$ & $O(N^2)$ & $O(C)$ & $O(C)$ \\
        \midrule
        WHITED & \textbf{Acc.} & \underline{0.7426} & {0.5448} & 0.2164 & 0.5125 & 0.4739 & \textbf{0.8284} & 0.2052 & 0.4291 & 0.7276 \\
        WHITED & Prec. & {0.7426} & {0.5107} & 0.1653 & 0.5125 & 0.4688 & \textbf{0.8522} & 0.0951 & 0.4175 & \underline{0.7866} \\
        WHITED & Recall & {0.7500} & {0.5194} & 0.1731 & 0.4620 & 0.4673 & \textbf{0.8165} & 0.1235 & 0.4265 & \underline{0.7735}\\
        WHITED & F1 & {0.7261} & {0.4887} & 0.1505 & 0.4484 & 0.4429 & \textbf{0.8138} & 0.0907 & 0.3934 & \underline{0.7583}\\
        \midrule
        BTS & Acc. & \underline{0.9848} & {0.9823} & 0.9710 & 0.9770 & 0.9766 & 0.9559 & 0.9726 & 0.9828 & \textbf{0.9854}\\
        BTS & Prec. & {0.7379} & {0.7429} & 0.6749 & 0.4925 & 0.4784 & 0.1838 & 0.6504 & \underline{0.8306} & \textbf{0.8813}\\
        BTS & Recall & \textbf{0.7067} & \underline{0.6920} & 0.1989 & 0.2336 & 0.2366 & 0.2704 & 0.3250 & 0.3231 & 0.4262 \\
        BTS & \textbf{F1} & \textbf{0.7219} & \underline{0.7166} & 0.3073 & 0.3169 & 0.3166 & 0.2189 & 0.4334 & 0.4652 & 0.5745\\
        \bottomrule
        \multicolumn{10}{l}{\footnotesize \textsuperscript{*}Due to the computational overhead of multi-label classification in Scikit-learn’s SVM implementation, we limit the maximum number of iterations for SVM training to 1000 for the BTS dataset.}
    \end{tabular}%
    }
    \label{table:classification}
\end{table*}

\textbf{Computational Complexity.} The computational complexity of the models varies significantly based on their training methodology. \texttt{DTW}, \texttt{SDTW} and \texttt{TS2Vec} exhibit quadratic scaling with the number of training samples due to their reliance on pairwise distance computations—\texttt{SDTW} for DTW calculations and \texttt{TS2Vec} for its contrastive learning framework. In contrast, \texttt{ResNet} scales linearly, as it is trained in a conventional supervised manner using class labels, without the need for pairwise comparisons. The most computationally efficient models in our study are \texttt{MOMENT} and \texttt{Chronos}, as their zero-shot encoders generate embeddings independently of the training sample size. Moreover, the subsequent SVM and NN classifiers can be trained on a subset of the available samples, further reducing computational demands. 

\textbf{Performance.} Performance evaluations indicate that \texttt{Chronos}, when paired with an SVM classifier, achieved state-of-the-art results on the WHITED dataset, even surpassing \texttt{DTW}. However, in the BTS dataset, where the F1 score serves as a more appropriate evaluation metric, \texttt{Chronos} underperformed relative to \texttt{DTW}. Notably, its combination with SVM exhibited significantly weaker performance, likely due to the classifier failing to converge within the imposed limit of 1,000 iterations—an issue attributed to the large scale of the BTS dataset. In contrast, replacing the SVM with an NN classifier substantially improved results, making \texttt{Chronos} the highest-performing approach after \texttt{DTW} and \texttt{SDTW}. 

\textbf{Representation.}
A key insight from our experiments is that the zero-shot embeddings generated by \texttt{Chronos} consistently outperform those learned through supervised contrastive training in \texttt{TS2Vec}, while \texttt{MOMENT} exhibits comparable performance. Unlike \texttt{TS2Vec}, which requires explicit training on each dataset to extract meaningful feature representations, \texttt{Chronos} and \texttt{MOMENT} generate embeddings directly from their pretrained models without additional fine-tuning. Despite the absence of dataset-specific adaptation, \texttt{Chronos} produces high-quality representations, as evidenced by its strong classification performance when paired with either an SVM or NN classifier. While \texttt{MOMENT} does not match \texttt{Chronos} in representation quality, it still performs on par with \texttt{TS2Vec}, despite the latter’s reliance on dataset-specific training. These findings underscore the effectiveness of TSFMs in capturing universal time-series representations without the need for additional fine-tuning.

\textbf{Classifier Head.} Classifier selection remains a critical factor in model performance, as different classifiers exhibit varying levels of effectiveness across datasets. For \texttt{TS2Vec}, \texttt{MOMENT}, and \texttt{Chronos}, our experiments reveal a consistent pattern: SVM performs better on WHITED, while NN performs better on BTS. This is expected, as NNs typically require larger datasets to generalize effectively, whereas SVMs can perform well even with limited training samples. Consequently, NNs tend to outperform SVMs on large datasets, while SVMs may be preferable when data availability is constrained. Overall, the absence of a universally optimal modeling approach aligns with the findings of \citep{xue2023utilizing}, which demonstrated that the most effective language models for forecasting varied across different buildings. These observations reinforce the necessity of task-specific and dataset-dependent model/classifier head selection to achieve optimal results.

\subsection{Summary}
\texttt{Chronos} demonstrates strong zero-shot representation capabilities, achieving competitive performance without requiring dataset-specific training. However, it still remains inferior to statistical models, in means of F1 for large datasets. \texttt{MOMENT} though inferior to \texttt{Chronos} has shown comparative performance to deep-learning based methods that require data-specific training (i.e., \texttt{TS2Vec}). Performance of TSFMs seems to vary considerably based on the classifier used, calling into attention that different classifiers should be chosen depending on the size of the dataset. Despite a possible performance gap for larger and imbalanced datasets (i.e., BTS), \texttt{Chronos}'s drastically lower computational complexity makes it a practical choice for resource-constrained environments, considering the trade-off between efficiency and predictive accuracy. 
\section{Robustness Analysis}
\label{sec:behavior}
This section evaluates model performance across varying conditions and metrics to identify potential failure modes when data characteristics shift. First, we examine how these models generalize to different time scales and forecasting horizons, assessing their adaptability to varying prediction requirements. Second, we analyze edge cases to determine whether they can accurately capture behavioral shifts, such as peak occurrences. Third, we evaluate their performance across alternative metrics beyond RMSE, providing insights into how different training objectives shape their forecasting characteristics. Finally, we assess their robustness under dynamic conditions, such as changes in occupancy patterns. While some of these scenarios may be infrequent, they remain statistically significant and critical for real-world applications.

\subsection{Datasets}

We specifically seek to test the performance of TSFMs with data subsets representing specific building conditions to understand whether there is significant performance degradation when factors governing the physical system change. To that end, we utilize two sets of data we collected for approximately six months from an experimental testbed with four occupants located in <Anonymous location>, which is released alongside the code for all experiments at <Anonymous Github Repo>.

\begin{figure}[t]
    \centering
    \begin{tabular}{ccc}
        \includegraphics[width=0.8\textwidth]{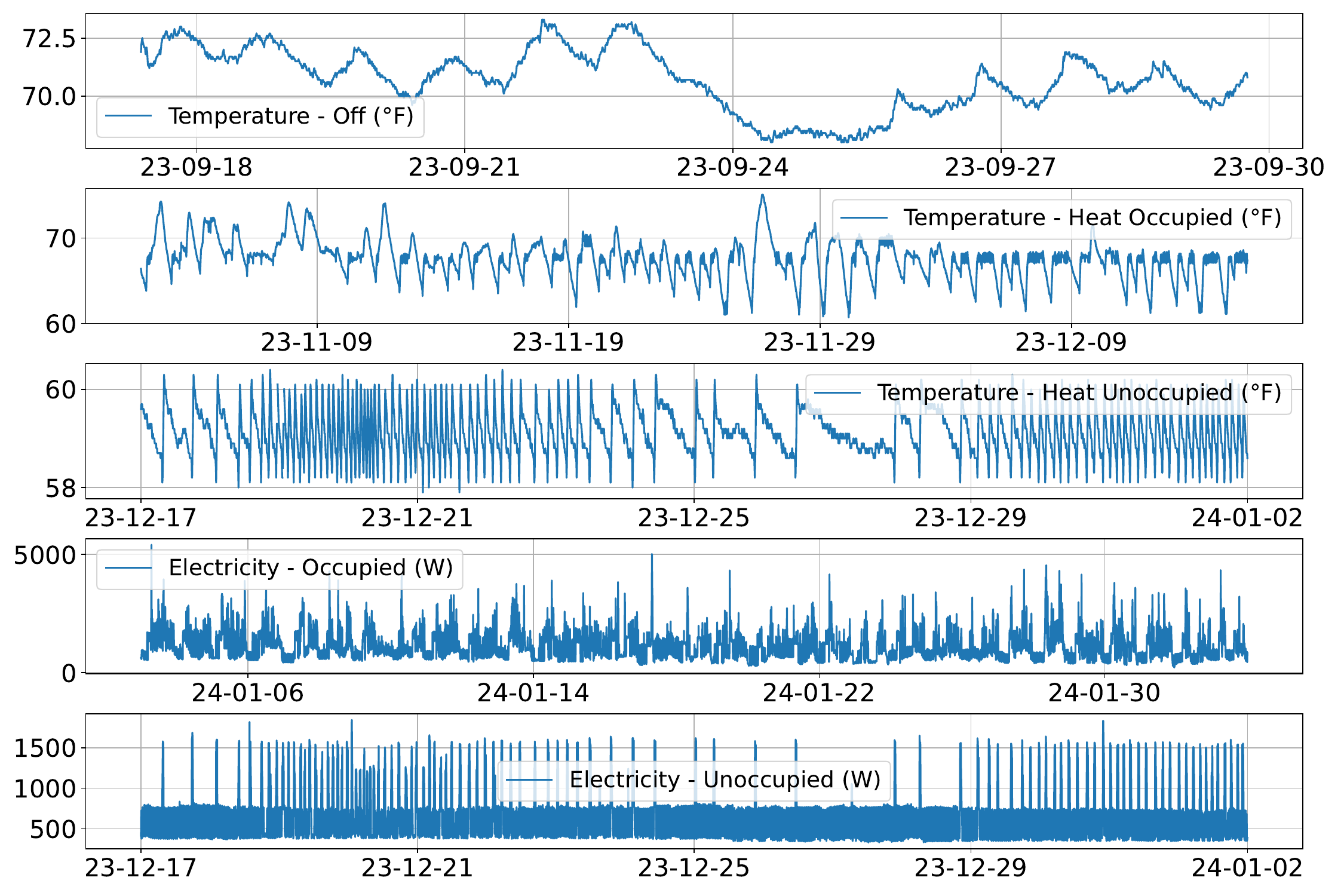} \\
    \end{tabular}
    \caption{Visualization of new empirical data across conditions}
    \label{fig:combined_vis}
\end{figure}

\noindent \textbf{Indoor Air Temperature.} An \textit{ecobee} thermostat\footnote{\url{https://www.ecobee.com/en-us/smart-thermostats/}} that takes measurements of 0.1°F resolution at five-minute intervals was utilized for data collection. Although this dataset comprises ten temperature sensors, our focus was exclusively on the thermostat temperature. From this dataset, we extracted three subsets based on distinct conditions (Figure \ref{fig:combined_vis}): \begin{itemize}
    \item \textbf{\emph{Off}:} This condition refers to the period when the HVAC system was off for an extended duration, specifically from September 17\textsuperscript{th} to September 29\textsuperscript{th}. During this time, although the house was occupied, the HVAC system was not operational due to warm outdoor conditions, resulting in a free-floating thermal environment. The purpose of this period is to facilitate univariate time-series forecasting, as the future values of the time-series are less dependent on other covariates due to the absence of heating/cooling input from HVAC.
    \item \textbf{\emph{Heat-Occupied}:} This condition covers the period when the HVAC system was set to heating mode and the house was occupied. We selected a duration during the winter, from November 2\textsuperscript{nd} to December 16\textsuperscript{th}, to ensure active heating. During this period, the house followed a schedule defined by the \textit{ecobee} thermostat, with setpoints of 62°F during sleep, 64°F when away, and 68°F when home. In this condition, we have a more variant temperature pattern as the setpoints change three times during the day in addition to the heat gains by occupants.
    \item  \textbf{\emph{Heat-Unoccupied}:} This condition pertains to the period when the HVAC system was in heating mode while the house was unoccupied, from December 17\textsuperscript{th} to January 2\textsuperscript{nd}. Throughout this period, the house maintained a stable heating setpoint of 60°F. This condition serves as a test for a more cyclic behavior since the setpoint stays the same for a long period.  
    
\end{itemize}

\noindent \textbf{Electricity.} A \textit{Sense} smart meter\footnote{\url{https://sense.com/}} was installed in our testbed to collect measurements of active power at five-minute intervals. Although data was collected over a period exceeding six months, we extracted two subsets based on distinct occupancy conditions (Figure \ref{fig:combined_vis}):
\begin{itemize}
    \item \textbf{Unoccupied:} We selected the same unoccupied period as the temperature data, from December 17\textsuperscript{th} to January 2\textsuperscript{nd}. This condition serves as a test for their ability in more cyclic patterns.
    \item  \textbf{Occupied:} This subset consists of data collected from January 3\textsuperscript{rd} to February 3\textsuperscript{rd} when the house was occupied. When occupied, there are more subtle peaks in the house and thus it is a good test of their performance on a more variant electricity input.

\end{itemize}

\subsection{Models}
We use the same TSFMs and baselines used for univariate forecasting, as explained in Section \ref{sec:univariate_models}.

\subsection{Experimental Setting}

We conducted experiments that vary across five dimensions:
\begin{enumerate}
    \item Data: \{Temperature, Electricity\}
    \item Condition: \{\emph{Off}, \emph{Heat-Occupied}, \emph{Heat-Unoccupied}\}, \{\emph{Occupied}, \emph{Unoccupied}\}
    \item Varying Context Duration $D$ (hours): \{24, 36, 48,96\}
    \item Varying Prediction Duration $P$ (hours): \{4, 6, 12, 24\}
    \item Varying Starting Points: We created 10 batches with different starting points for each context-horizon pair by dividing the available data into equal intervals, ensuring each model uses the same set of starting points.
\end{enumerate}

\subsection{Metrics}

Existing work mainly uses metrics that measure magnitude differences between forecasts and ground truth such as RMSE, MASE, and MSE. While these metrics are useful for understanding the quality of predictions, they are intolerant to phase differences, resulting in considerably large values in such cases. Thus, we extend measurement rules to include metrics that quantify the shape similarity between the predictions and the ground truth time-series. In that regard, we choose Peak Recall Rate (PRR) and Dynamic Time Warping (DTW) in addition to RMSE to characterize predictions.

\noindent \textbf{Peak Recall Rate (PRR).}
Peak patterns in time-series refer to the significant increases in the magnitude of measured variables followed by corresponding decreases, which are common features in air temperature and electricity building data \citep{fan2014development, braun1990reducing}. Correctly forecasting these peaks offers substantial benefits, including enhanced energy efficiency, cost savings, and improved occupant comfort \citep{braun1990reducing}. For instance, building control systems can make proactive energy adjustments to balance grid load and participate in demand response programs for financial incentives. 

Therefore, we propose to use the peak recall rate (PRR) as the metric to evaluate models' ability to forecast peaks and define PRR as follows:

Let $s$ be the ground-truth time-series and $\hat{s}$ be the forecasted time-series. Let $\mathbb{P}$ be the set of peak indices in $s$, and $\mathbb{\hat{P}}$ be the set of peak indices in $\hat{s}$. We calculate the peak recall rate as:

\begin{equation}
    \text{PRR} = \frac{1}{|\mathbb{P}|} |\{ i \in \mathbb{P} | \exists j \in \mathbb{\hat{P}} \text{ s.t. } |i - j| \leq \tau \text{ and } |s_i - s_j| \leq \varphi \}|
\end{equation}

Where $\tau$ and $\varphi$ are two empirically chosen thresholds to ensure the closeness of the true peak and the forecasted peak in time and magnitude. Typically, we set $\tau = H/20$ ($H$ is the prediction length) and $\varphi = \sigma$ ($\sigma$ is the standard deviation of $s$) to ensure the two thresholds are adaptive to different samples. Using PRR as the metric can shed light on the fidelity of peak patterns in building data forecasting.

\noindent \textbf{Dynamic Time Warping Index (DTW).} DTW \citep{muller2007dynamic} measures the similarity between two time-series by finding the optimal alignment that minimizes the cumulative distance between them. Unlike traditional distance metrics that require one-to-one comparisons, DTW allows for elastic shifts in the time dimension, effectively aligning sequences that may be out of phase.

By warping the time axis, DTW aligns each point in one series with one or more points in another, minimizing the overall distance. This metric tolerates the case where forecast models capture the correct overall trend but differ in timing or pace, providing a robust measure of accuracy that accounts for temporal distortions. The flexibility of DTW makes it useful to recognize underlying similarities despite temporal shifts.

\subsection{Results}

This analysis aims to compare the forecasting performance of various TSFMs using the subsets of data defined above. We have resampled the data to keep $H < 64$, as most of these models are optimized for a prediction horizon ($H$) of 64 steps.  

\begin{figure*}[h]
    \centering
    % \begin{tabular}{c}
    \hspace{2.9mm}
        \includegraphics[width=0.92\textwidth]{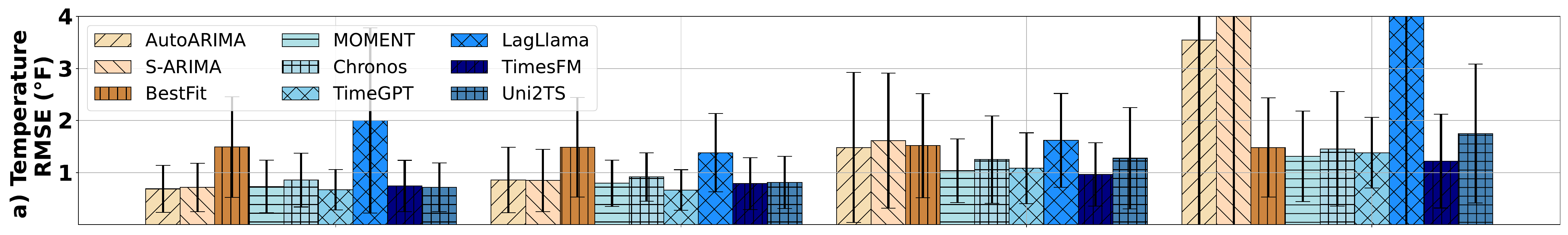 } \\
        %(a) Temperature \\
        \vspace{-1.2mm}
        \includegraphics[width=0.94\textwidth]{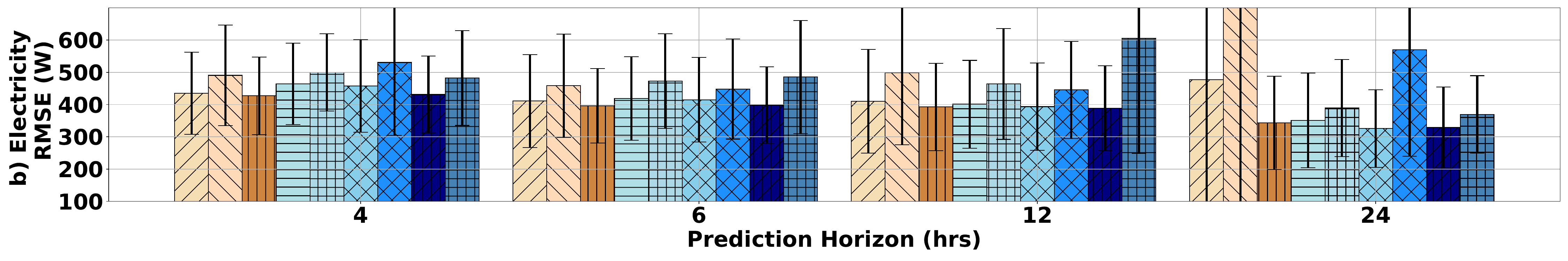}  \\
        %(b) Electricity \\
    % \end{tabular}
    \caption{RMSE with varying prediction horizon ($P$). The context durations ($D$) are averaged}
    \label{fig:vary_hrz}
    \vspace{-4mm}
\end{figure*}

\textbf{Varying Prediction Duration ($P$).} Given that our models only use data available in the context window as input, we expect forecasting errors to increase with longer prediction durations due to higher uncertainty and complex dynamics. To evaluate this hypothesis, Figure \ref{fig:vary_hrz} shows the forecasting performance, given in RMSE of Fahrenheit and Watts, with varying prediction horizons in hours, averaged over context length. The best TSFMs achieve RMSEs increasing with longer prediction horizons ($P$) while being around 1°F overall. \texttt{AutoARIMA} and \texttt{S-ARIMA} show competitive performance with TSFMs for shorter horizons ($P=4$ and $6$), but their errors increase for longer horizons ($P=12$ and $24$). Besides, \texttt{LagLlama} displays higher RMSEs across most horizons, indicating less accurate forecasts compared to all other data-driven and statistical-based models. In summary, most TSFMs and statistical models have less than 1°F error in prediction for shorter horizons ($P=4$ and $6$), and statistical models become inferior to TSFMs for longer horizons ($P=12$ and $24$).

When analyzing results from electricity use predictions (Figure \ref{fig:vary_hrz} (b)), TSFMs achieve around 400 W RMSE consistently, and their performance further improves for $P=24$. \texttt{BestFit} shows competitive performance compared to the best TSFMs (\texttt{TimesFM} and \texttt{TimeGPT}) across all horizons. Similar to the temperature data, \texttt{AutoARIMA} and \texttt{S-ARIMA} demonstrate competitive performance for shorter horizons but are outperformed by \texttt{MOMENT}, \texttt{TimeGPT}, and \texttt{TimesFM} for longer horizons. \texttt{MOMENT}, \texttt{Chronos}, \texttt{LagLlama}, \texttt{TimesFM}, and \texttt{Uni2TS} models show varied performance, with none consistently outperforming the statistical-based models. Notably, only \texttt{TimesFM} shows competitive performance across all horizons.

Overall, the results show that TSFMs achieve lower forecasting error for longer prediction horizons, aligned with our initial hypothesis. For instance, the scenario of $P=24$ with $D=24$ is challenging, where prediction becomes more dependent on model's familiarity with the phenomenon since the context window is relatively short. Besides, the ability of TSFMs to interpolate future data with learned dynamics might not be as prominent as anticipated, indicated by the fact that \texttt{BestFit} is competitive with the top TSFMs in electricity forecasting.

%ARIMA-based models (\texttt{AutoARIMA} and \texttt{S-ARIMA}) are effective for shorter horizons, but their performance diminishes as the horizon extends.  %(\texttt{AutoARIMA} produces RMSEs of 3.549 and 476.898 on temperature and electricity data for the 24-hour horizon, respectively). 

\begin{figure*}[h]
    \centering
    % \begin{tabular}{c}
    \hspace{2.9mm}
        \includegraphics[width=0.92\textwidth]{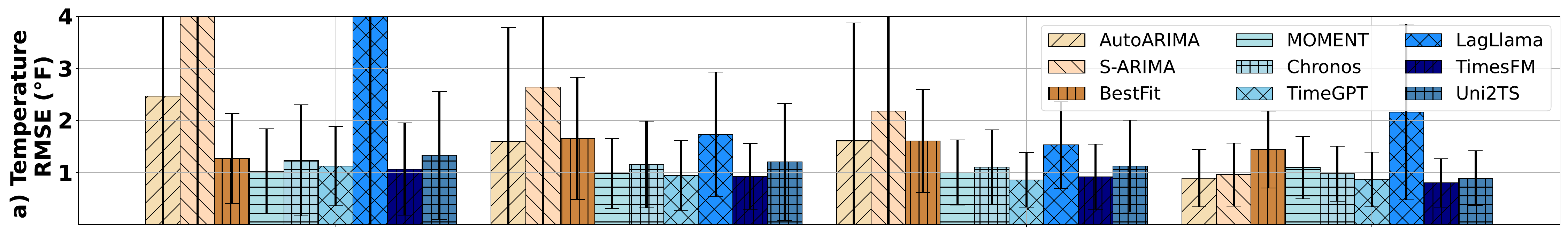 } \\
        %(a) Temperature \\
        \vspace{-1.2mm}
        \includegraphics[width=0.94\textwidth]{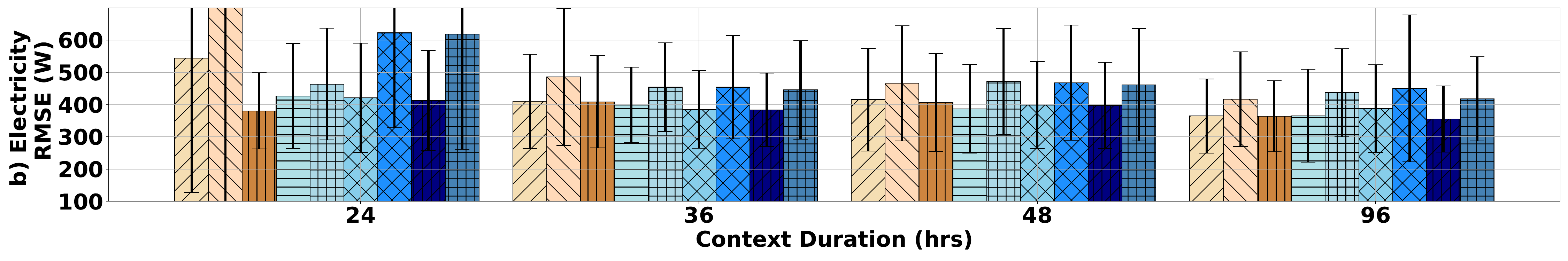} \\
        %(b) Electricity \\
    % \end{tabular}
    \caption{RMSE with varying context duration ($D$). The prediction horizons ($P$) are averaged}
    \label{fig:vary_con}
\end{figure*}

\textbf{Varying Context Duration ($D$).} We hypothesize that the length of the context window and forecasting error are inversely correlated. This implies that for a fixed prediction horizon, the actual time-series data preceding the prediction is more informative than data the model learned from other similar time-series. As a result, statistical models that leverage the actual preceding data may outperform TSFMs in certain cases.

To analyze this hypothesis, we present the results in Figure \ref{fig:vary_con}, which shows the forecasting performance with varying context windows in hours, averaged over the prediction horizon. Figure \ref{fig:vary_con} (a) demonstrates that except \texttt{LagLlama}, all other foundation models achieve better performance than statistical models when $D=24$, $36$, $48$. However, the condition changes when $D=96$. \texttt{AutoARIMA} and \texttt{S-ARIMA} can achieve performance comparable to that of TSFMs. In predictions of electricity (Figure \ref{fig:vary_con} (b), \texttt{BestFit} achieves strong performance across all context duration. It is comparable to the best TSFMs (\texttt{TimesFM} and \texttt{TimeGPT}) consistently, and it defeats three TSFMs in all cases. \texttt{AutoARIMA} and \texttt{S-ARIMA} are less effective for shorter context windows but remain competitive when $D=36$, $48$, and $96$ hours.

The results support our hypothesis that the length of the context window and forecasting error are inversely correlated. Besides, TSFMs exhibit varying performance across different context windows and prediction horizons, with no single model consistently outperforming statistical-based models in all scenarios. Statistical models struggle with short context durations ($D=24$) due to insufficient data for optimal parameter estimation. However, as the context duration increases ($D \geq 36$), statistical models effectively leverage preceding data, becoming more robust and occasionally surpassing TSFMs.

\begin{figure}[t]
    \centering
        \includegraphics[width=0.96\textwidth]{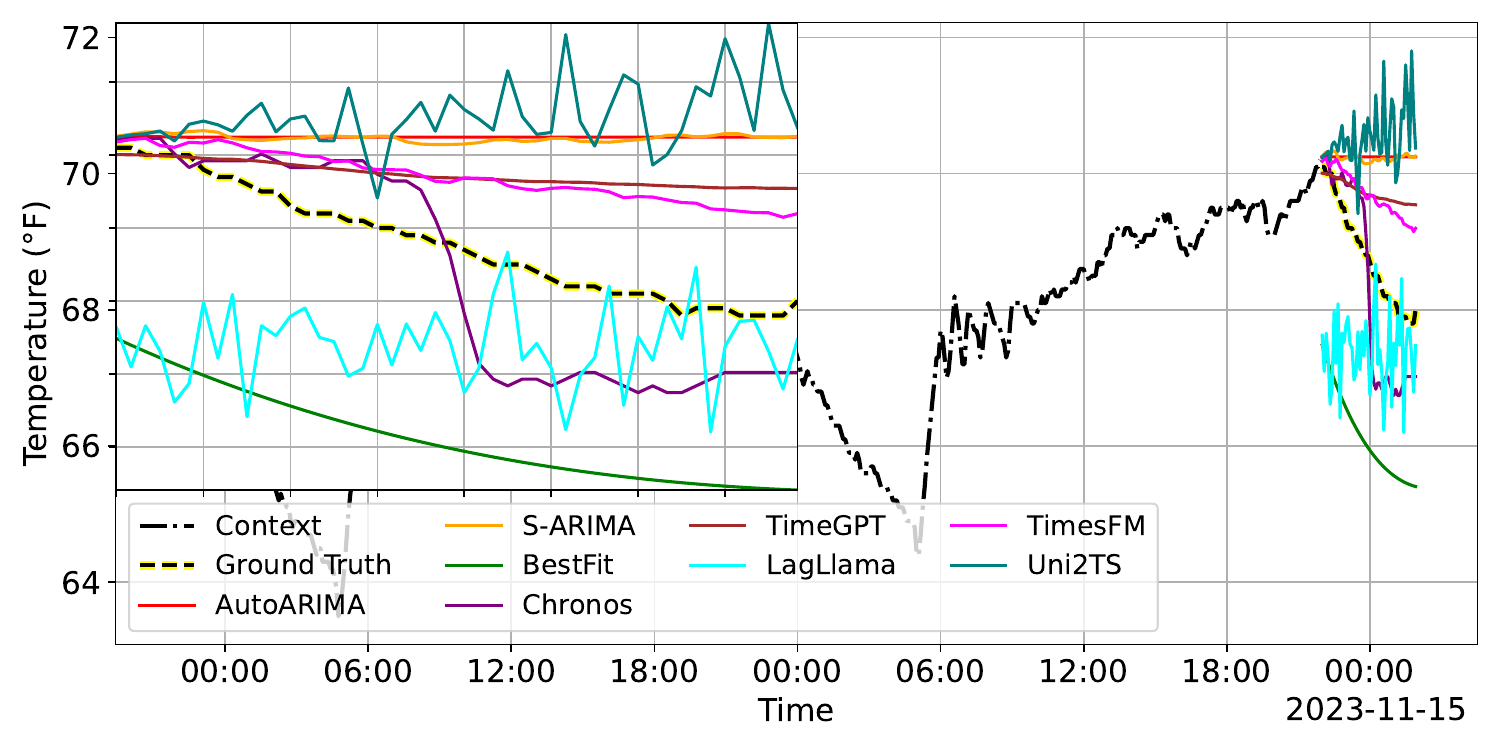} \\
        \small{a) Temperature peak during a \emph{heat-occupied} condition} \\
        \includegraphics[width=0.96\textwidth]{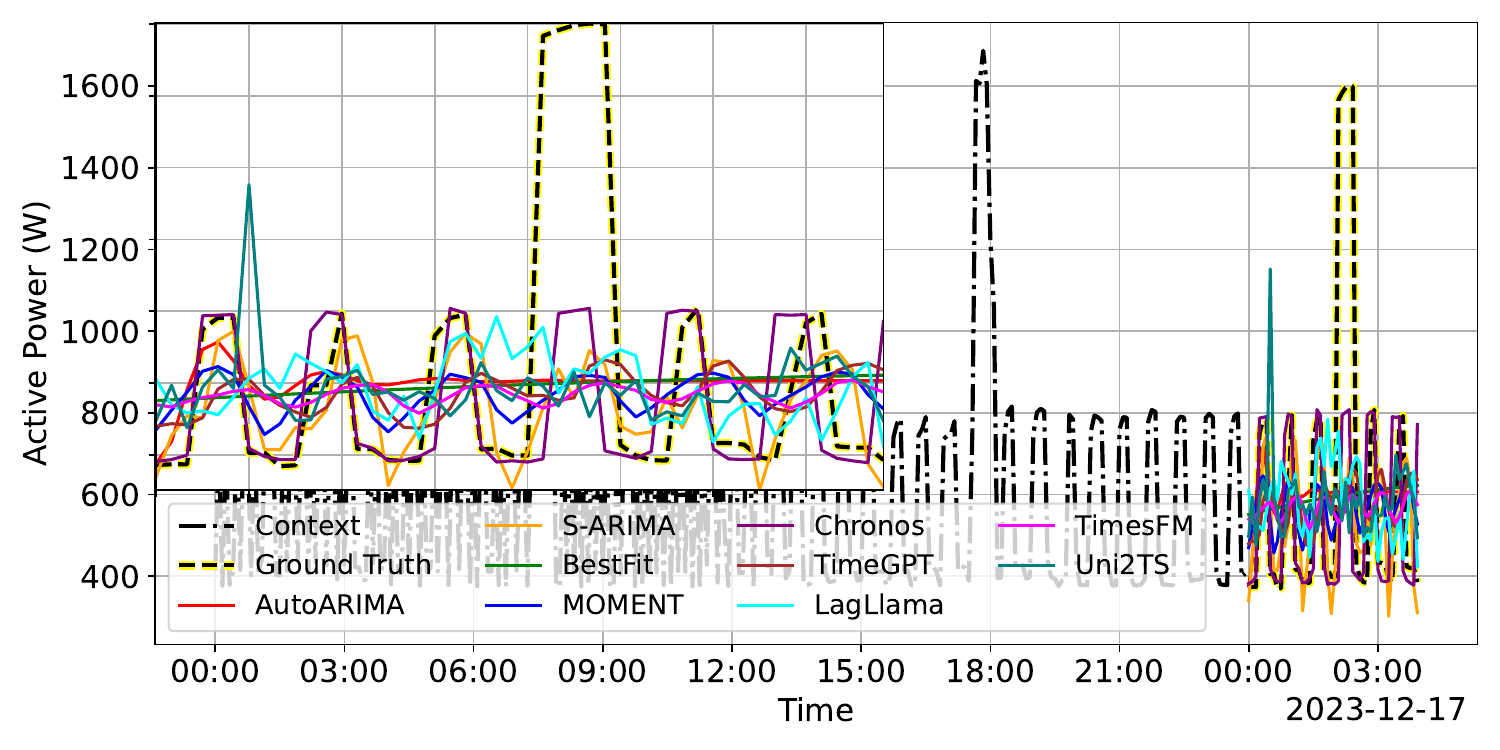 } \\
        \small{b) Electricity during an \emph{unoccupied} condition} \\
    \caption{Temperature and electricity predictions during heat-occupied and unoccupied conditions, illustrated with detailed views of key periods}
    \label{fig:corner}
\end{figure}

\textbf{Corner Cases.} In most studies, the performance of predictive models has been evaluated using the RMSE. However, RMSE alone does not provide a complete understanding of \textit{how} these models make predictions. To address this, we first illustrate the performance of these models in two corner cases. Figure \ref{fig:corner}(a) presents a scenario where the temperature reaches a peak and then starts to decline. This is particularly challenging for univariate models, which lack information on external factors and rely solely on previous temporal patterns present in the forecasted variable. Despite this, \texttt{Chronos}, \texttt{TimesFM}, and \texttt{TimeGPT} successfully anticipated this decline. In contrast, Figure \ref{fig:corner} (b) shows a case with a significant peak in electricity consumption within the prediction horizon, which is also observed in the context window. None of the models accurately predicted this peak. Nevertheless, only \texttt{Chronos} and \texttt{S-ARIMA} captured a similar cyclic pattern, while \texttt{MOMENT} recognized the cycles but failed to predict the correct magnitude. A key issue here is that RMSE might inflate errors for models that misplace peaks, even if they capture the overall trend, compared to models that average out the predictions. From these corner cases, we derive two main insights: (1) Univariate predictions struggle to be accurate in situations influenced by external factors (e.g., outdoor air temperature, heating gain). Therefore, models that can incorporate covariates are essential, particularly for temperature predictions. (2) The shape of the predictions is as important as the magnitude differences over time. Consequently, we should also assess the comparative performance of these models using a metric that evaluates pattern similarity between the predictions and the ground truth.

\begin{figure}[t]
% \vspace{-5mm}
    \centering
        \includegraphics[width=0.9\textwidth]{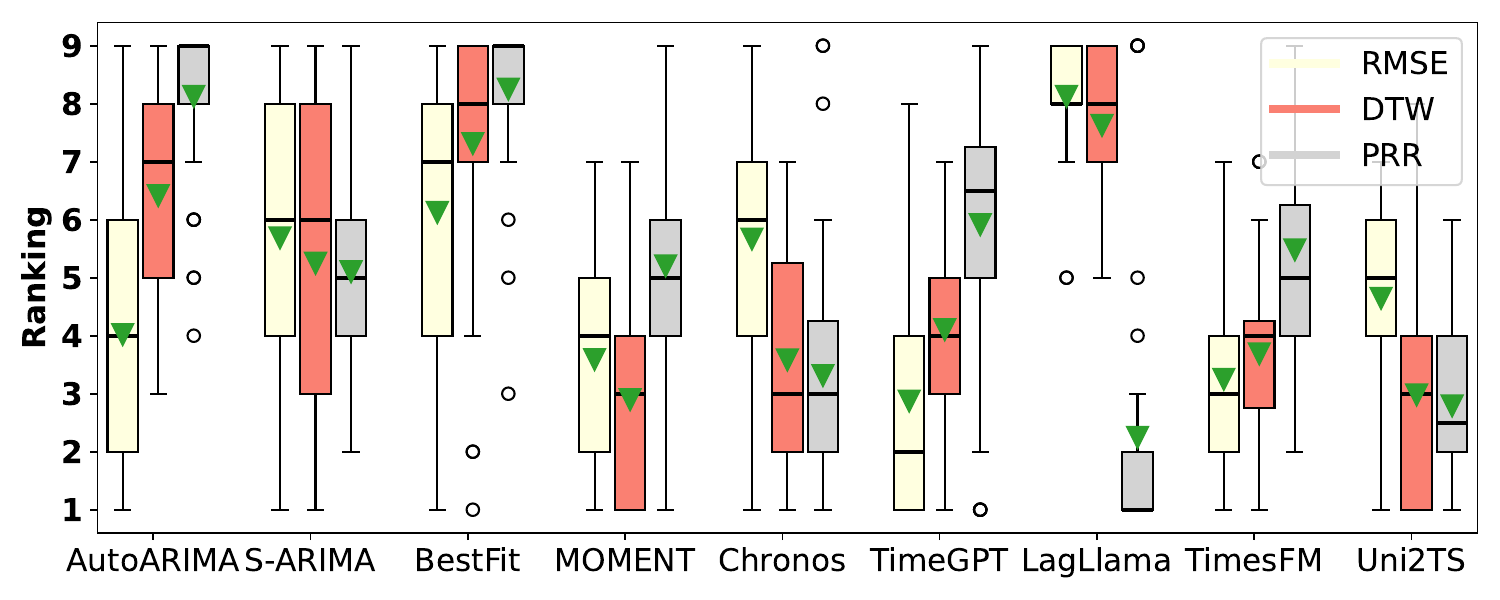}  \\
        \small{a) Distribution of Rankings for Temperature Data} \\
        \includegraphics[width=0.9\textwidth]{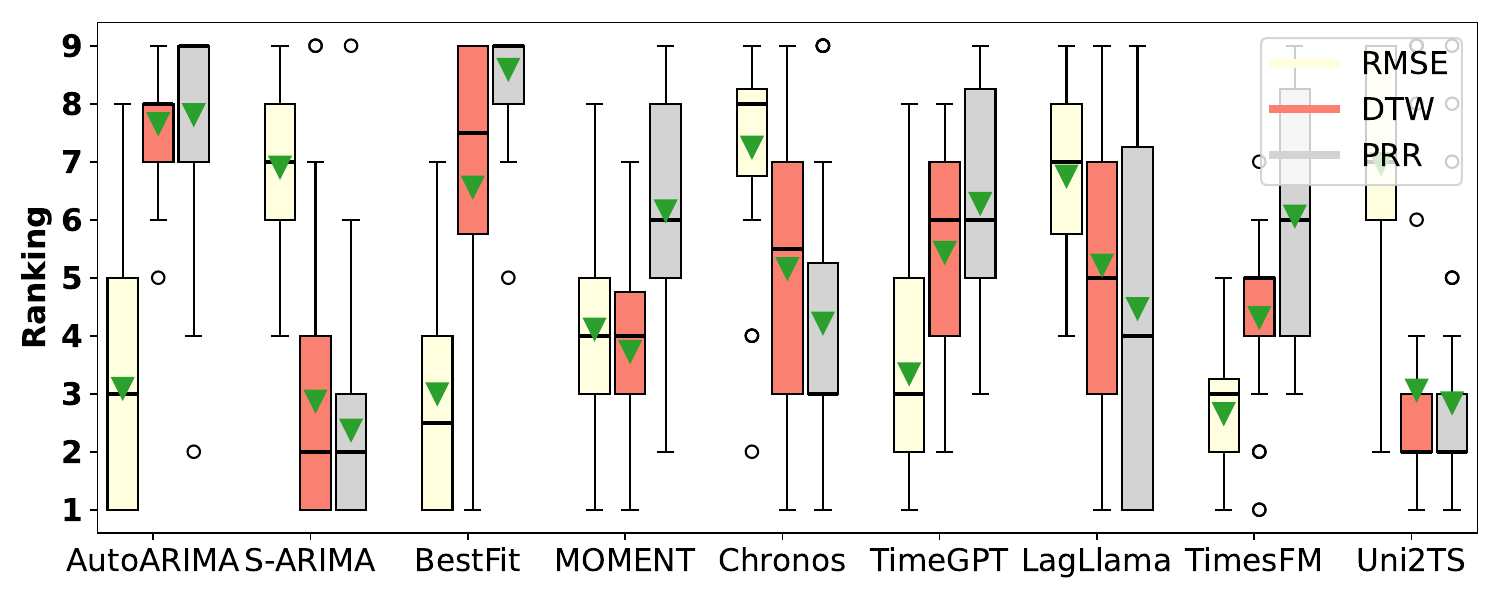} \\
        \small{b) Distribution of Rankings for Electricity Data} \\
    \caption{Model rankings across various metrics for temperature (a) and electricity (b) data. The green triangle denotes the mean, while the black line is the median}
    \label{fig:combined}
\end{figure}

\textbf{Effect of different metrics: RMSE, DTW, PRR.} To demonstrate the difference in comparative performance across models, we compute the ranking of each model at each condition based on different metrics. Figure \ref{fig:combined} shows the distribution of those rankings for each model for temperature and electricity data. 

For temperature data (Figure \ref{fig:combined} (a)), when evaluated by the RMSE metric, \texttt{MOMENT}, \texttt{TimeGPT}, and \texttt{TimesFM} are the best three TSFMs, followed by the statistical model \texttt{AutoARIMA}. Among all TSFMs, \texttt{Chronos}, \texttt{LagLlama}, and \texttt{Uni2TS} are the three lowest-ranked models when evaluating with RMSE. However, the ranking changes completely when DTW and PRR are used: \texttt{MOMENT}, \texttt{Chronos}, and \texttt{Uni2TS} are the top three models ranked by DTW. This discrepancy in ranking shows that there are models better at matching the pattern similarity of the ground truth than predicting the magnitude. More interestingly, when evaluated by the PRR metric, \texttt{LagLlama}, \texttt{Uni2TS}, and \texttt{Chronos} become the top three models, while they are the worst three TSFMs evaluated by RMSE. This result shows that RMSE is biased towards models that predict general tendency and suppress peak values, where the misalignment of the peaks could produce large RMSE.

We could also observe a similar pattern in model performance in electricity data as shown in Figure \ref{fig:combined} (b). When evaluated by RMSE, \texttt{MOMENT}, \texttt{TimeGPT}, and \texttt{TimesFM} are the best three TSFMs, while \texttt{LagLlama}, \texttt{Uni2TS}, and \texttt{Chronos} are the worst three. However, when evaluated by DTW, \texttt{LagLlama}, \texttt{Uni2TS}, and \texttt{TimesFM} are the best three TSFMs. And likewise, \texttt{LagLlama}, \texttt{Uni2TS}, and \texttt{Chronos} are the top three evaluated by PRR. A similar decrease also happens to statistical models such as \texttt{AutoARIMA} and \texttt{BestFit}, and \texttt{S-ARIMA} demonstrates an increase in performance when DTW and PRR are used. Both experiments in electricity and temperature data demonstrate that the winners evaluated by RMSE may not capture the full properties of the time-series.

We conjecture that the inconsistency in models' behaviors is related to the objective metric used in training. \texttt{LagLlama}, \texttt{Uni2TS}, and \texttt{Chronos} use next-token or masked-token prediction in training, where the Cross-Entropy (CE) or Negative Log-Likelihood (NLL) loss is the objective metric. Using CE or NLL loss maximizes the likelihood of the ground truth, which encourages the model to focus on the overall distribution and temporal dependencies. Specifically, by modeling the conditional probability of forecasting given the context, the model can capture the temporal dependencies for preserving the overall shape and patterns in the data. On the other hand, \texttt{MOMENT} and \texttt{TimesFM} use the MSE loss as the objective, which is sensitive to outliers and leads to smoother predictions as the metric penalizes large deviations. Different choices of objective function direct the models to treat peaks and shape patterns differently. Models trained with NLL or CE loss may still produce sharper adjustments when the probabilities are significantly wrong, while those trained with MSE loss would smooth the error magnitudes.

In summary, while we are not making conclusive claims, we believe it is noteworthy that RMSE is only one metric for evaluating performance and that we should utilize other metrics in understanding TSFM's forecasting abilities. These differences in model ranking reveal that RMSE can mask the abilities to preserve shapes and predict peak patterns of TSFMs, which are important features in building control systems.

\textbf{Varying Conditions.} Our analysis indicated potential differences in model predictive performance with varying durations and evaluation metrics, necessitating a more contextualized examination of data patterns in building physics. For electricity, the \emph{occupied} condition shows more variable behavior, while \emph{unoccupied} is more cyclic. In heating data, \emph{heat-occupied} is more variable, whereas \emph{heat-unoccupied} maintains a cyclic pattern. The \emph{off} condition demonstrates a case where the univariate time-series is less dependent on other covariates. Examples can be found in Figure \ref{fig:combined_vis}.
 
\begin{figure}%[h]
    \centering
    % \begin{tabular}{c}
        % \includegraphics[width=0.47\textwidth]{figures/comparison_condition.pdf } \\
        % (a) Temperature \\
        \hspace{2.4mm}
        \includegraphics[width=0.9\textwidth]{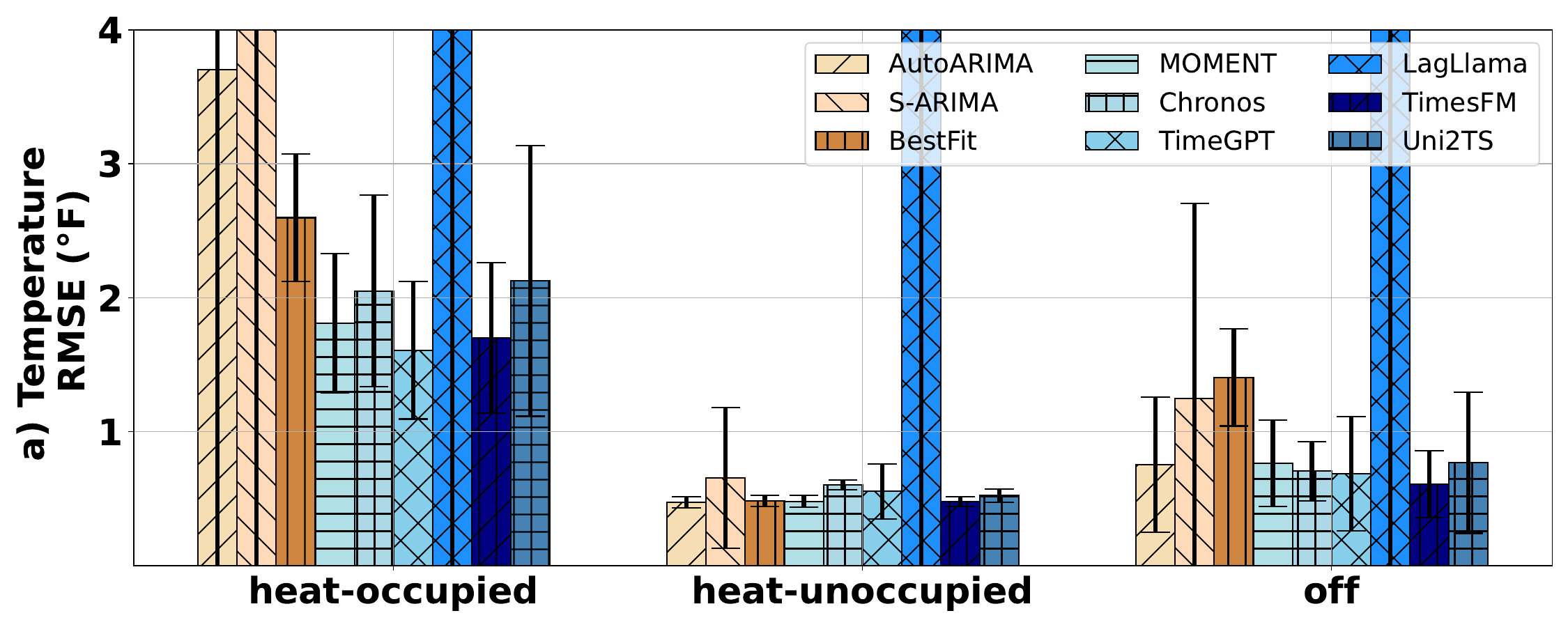 } \\
        \vspace{-1mm}
        % \small{(a) Temperature} \\
        \includegraphics[width=0.9\textwidth]{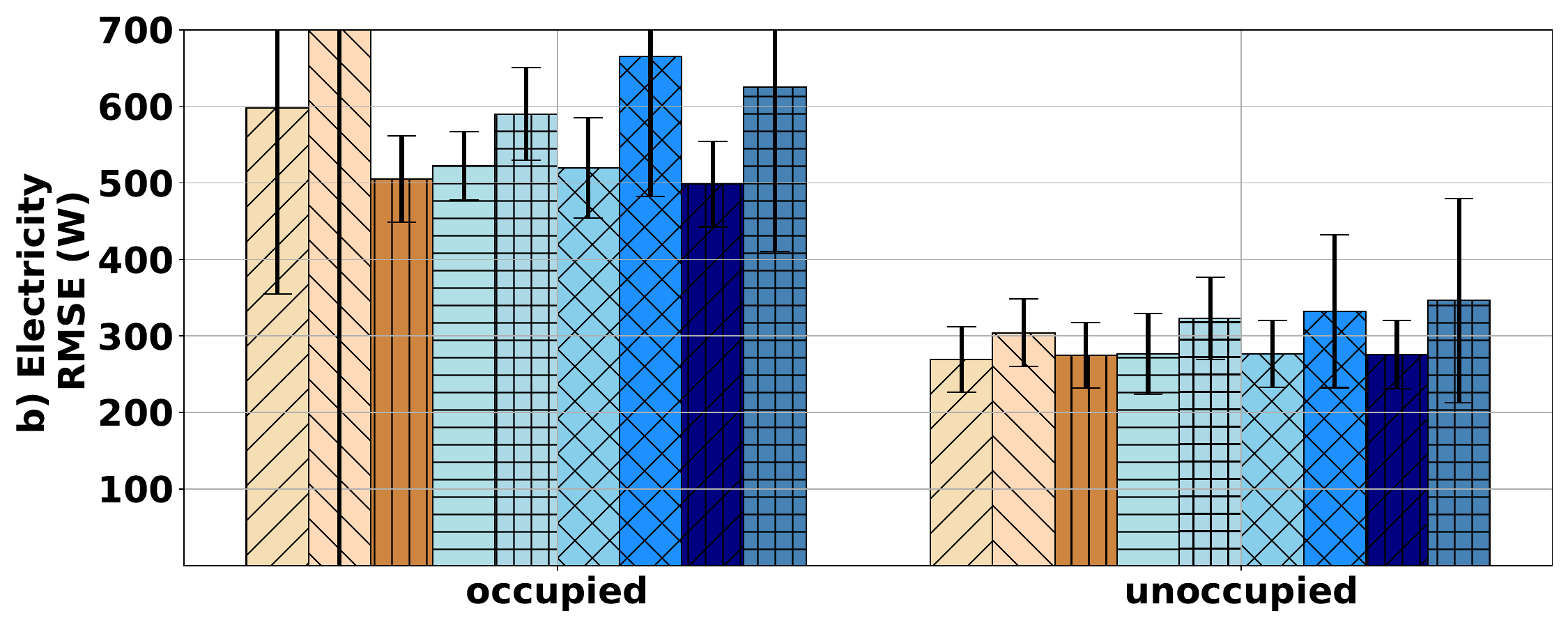 } \\
        \vspace{-1mm}
        % \small{(b) Electricity} \\
    % \end{tabular}
    \vspace{-1mm}
    \caption{RMSE on temperature and electricity data of various building conditions}
    \vspace{-6mm}
    \label{fig:vary_cond}
\end{figure}

Figure \ref{fig:vary_cond} demonstrates the distribution of RMSE values across models for different conditions each dataset possesses. Figure \ref{fig:vary_cond} (a) shows that \texttt{AutoARIMA} performs much better when the pattern is more cyclic and univariate (\emph{heat-unoccupied} and \emph{off}), while the performance drops for \emph{heat-occupied}. \texttt{BestFit} is only reliable on \emph{off} conditions. On the other hand, the performance variations of TSFMs are relatively small across \emph{heat-occupied}, \emph{heat-unoccupied}, and \emph{off}, with the only exception that \texttt{LagLlama}'s error increases on \emph{off}. Overall, statistical models are more reliable on more cyclic and univariate conditions like \emph{heat-unoccupied} and \emph{off} conditions, respectively. Besides, Figure \ref{fig:vary_cond} (b) displays that the contextualized dynamics of electricity data forecasting. In \emph{occupied} conditions, \texttt{BestFit} achieves demonstrates almost the same performance as the best TSFM (i.e. \texttt{TimesFM}). Similarly, \texttt{AutoARIMA} achieves even better performance than other TSFMs in means of ranking for \emph{unoccupied} conditions. Thus, for electricity, TSFMs demonstrate a marginal performance difference if not worse than statistical models. 

In summary, the differences in behaviors of statistical models indicate that statistical models are better than foundation models in more cyclic and simpler pattern predictions, while foundation models are more capable in complicated pattern forecasting. However, in most cases we observe the difference to be marginal except the \emph{heat-occupied} case.

\subsection{Summary}
In summary, broadening evaluation metrics to include DTW and PRR offers a deeper layer of analysis. While TSFMs like \texttt{TimesFM} and \texttt{TimeGPT} excel under RMSE assessments, models such as \texttt{S-ARIMA}, \texttt{Chronos}, and \texttt{Uni2TS} demonstrate superior pattern recognition capabilities, crucial for applications requiring precise pattern detection. The enhanced performance under DTW and PRR evaluations of models trained with CE or NLL losses suggests that the training objective enriches different aspects of their predictive outputs.

The impact of external conditions like HVAC operational modes and occupancy also delineates the strengths and limitations of these models in varying scenarios. \texttt{AutoARIMA} and \texttt{TimeGPT} excel in predictable, cyclic conditions such as unoccupied heating scenarios. In contrast, TSFMs are more proficient in managing complex, occupied environments, although their effectiveness is constrained, with prediction errors around 2°F in scenarios like an occupied house with frequent setpoint changes.

These observations emphasize the need for tailored evaluation of TSFMs, considering specific application requirements and operational conditions to strike a balance between accuracy, pattern fidelity, and dynamic response capabilities for efficient energy management and climate control. Despite considerable focus on TSFM forecasting capabilities, we observe that their performance does not significantly exceed that of simpler univariate statistical models.

\section{Discussion}
\label{sec:discussion}

\subsection{Summary}

We arrived at the following findings through the evaluation of TSFMs on forecasting tasks for buildings:
\begin{itemize}[leftmargin=*]
    \item \textbf{Generalizability.} (1) Dataset-level familiarity: By applying pretrained TSFMs to electricity datasets, we observe that TSFMs are only superior to statistical models on previously seen data. Some FMs can better generalize to unseen electricity datasets only with marginal difference compared to statistical models. (2) Modality-level familiarity: When evaluated on a large-scale dataset which TSFMs have limited familiarity with (i.e. indoor air temperature data) across various houses and seasons, TSFMs perform better for longer durations while statistical models outperform TSFMs for shorter durations. Besides, AutoARIMA remains competitive overall, nearly matching the performance of the top TSFMs on unseen datasets for seen sensor modalities. 
    \item \textbf{Covariate-utilization.} The inclusion of covariates generally did not enhance TSFM performance, likely due to their training being focused on univariate forecasting tasks. Compared to traditional methods, which are designed for one-step-ahead predictions and subsequently used for autoregressive forecasting, TSFMs showed inferior performance with or without covariates.
    \item \textbf{Representation.} For classification tasks, we leveraged \texttt{Chronos}'s and \texttt{MOMENT}'s capabilities to produce generalizable embeddings without task-specific fine-tuning. While, in some cases, inferior to non-parametric optimization models, \texttt{Chronos} and \texttt{MOMENT} delivered superior results compared to other representation learning techniques and deep-learning-based classifiers. This underscores the potential of TSFMs to derive valuable and generalizable representations without any dataset-specific training, attributed to the extensive corpus on which they were trained.
    \item \textbf{Stability.} Each TSFM's performance varies on different evaluation metrics. Some excel at preserving patterns and peak fidelity, as seen in DTW and peak recall assessments, but score poorly on Root Mean Squared Error (RMSE) evaluations. This variance is linked to the training objectives: models trained to mimic the true data distribution better preserve shapes and peaks, whereas those trained with Mean Squared Error (MSE) prioritize reducing magnitude errors. Therefore, it is necessary to pick the right model that matches one's task needs.
    \item \textbf{Adaptability.} Reflecting the confounding factors existing in buildings—specifically, the effects of occupancy and HVAC control mechanisms, we conducted a contextualized analysis for varying conditions. We observe some TSFMs are comparatively more powerful under occupied conditions (which have a more variant nature), while others are better under unoccupied conditions (i.e. cyclic patterns). Temperature and electricity data under unoccupied (i.e. constant setpoint) conditions often exhibit clear cyclic patterns where statistical models generally outperform TSFMs.
\end{itemize}

\subsection{Future Directions}

Our investigation in this paper is motivated by the belief that TSFMs, trained on a vast array of time-series data from diverse sources, might be able to generalize to previously unseen settings and datasets. The results presented in the previous section, while inconclusive due to their limited scope, suggest that this belief may be false. Beyond the obtained empirical insights into the current TSFM in building energy management tasks, we also highlight several areas that need further investigation and state key features that are needed by TSFMs.

\textbf{Towards Context-Aware and Task-Agnostic Time-Series Foundation Models:} Our analysis highlights critical challenges in applying TSFMs to real-world scenarios, particularly their inability to incorporate external context and handle open-ended task specifications. While models like \texttt{UniTime} \citep{liu_unitime_2024}, \texttt{Time-MQA} \citep{kong2025time}, and \texttt{TimeLLM} \citep{jin_time-llm_2024} incorporate some contextual cues, they are not designed for zero-shot prediction. Current zero-shot TSFMs rely exclusively on time-series data, missing auxiliary information that could significantly enhance their predictive capabilities. Unlike language models, which inherently benefit from syntactic and semantic context, time-series data lack such structural cues, making pattern inference more challenging. Although initial attempts have been made to integrate context into time-series forecasting \citep{quanreimagining}, the results remain suboptimal.

Moreover, the heterogeneity of time-series data—arising from diverse physical processes, sensor characteristics, and measurement resolutions—introduces complexities not typically encountered in text data. While language models achieve strong generalization by leveraging massive datasets, TSFMs struggle with out-of-domain adaptation due to their inability to incorporate auxiliary contextual information. Addressing this limitation requires architectural innovations that enable TSFMs to dynamically integrate contextual metadata and flexibly adapt to various task definitions.

\textbf{Language for Context and Task Definition:} To make TSFMs more versatile and practical, we propose two key enhancements: (1) incorporating a natural language channel for context integration and (2) enabling task-agnostic reasoning. 

First, TSFMs should evolve beyond fixed-task forecasting models and support a broader range of tasks, including classification, anomaly detection, and open-ended reasoning. Current TSFMs require retraining or fine-tuning for each new task, limiting their scalability in real-world applications where task definitions are fluid and training data are often scarce. Recent work \citep{kong2025time} has demonstrated the potential of fine-tuning language models for question answering for open-ended time series tasks, indicating a paradigm shift from fixated set of tasks for time series analysis. TSFMs should follow a similar paradigm to be \textit{truly} foundation models.

Second, TSFMs should integrate contextual metadata through a natural language channel, allowing users to specify conditions such as operational settings, weather conditions, occupancy levels, and system constraints. Context plays a crucial role in time-series predictions—particularly in domains like energy management, where external factors heavily influence system behavior. The absence of structured mechanisms to incorporate such metadata weakens TSFMs’ adaptability to real-world conditions. By enabling models to process both raw time-series data and auxiliary metadata in natural language form, TSFMs could significantly improve their predictive accuracy and interpretability.

\textbf{Proposed Advancements in TSFM Architecture:} To realize these capabilities, future TSFMs should integrate the following advancements:
\begin{itemize}
    \item Dual-input capability: TSFMs should be designed to accept both time-series data and natural language inputs, enabling them to incorporate structured and unstructured information for more context-aware predictions.
    \item Task-agnostic output mechanisms: Instead of relying on fixed task-specific heads, TSFMs should either employ pre-trained modular heads (as seen in \texttt{MOMENT} \citep{goswami_moment_2024}) or generate outputs in natural language, allowing them to unify multiple analytical tasks without retraining.
\end{itemize}

These enhancements would transform TSFMs from narrow-task predictors into flexible, context-aware analytical tools. By incorporating a language channel, TSFMs could bridge the gap between predictive modeling and real-world decision-making, improving usability and enabling more interactive, user-driven time-series analysis.

\section{Conclusions}
\label{sec:conclusions}
TSFMs claim to be generalizable across different domains for various tasks, leveraging the vast time-series data they have been trained on. Our findings indicate a marginal performance difference between TSFMs and statistical models on previously unseen datasets for univariate forecasting tasks. In particular, within domains such as indoor air temperature, where TSFMs lack prior exposure to familiar datasets, their performance varies but remains closely aligned with that of statistical models. Notably, incorporating covariates did not lead to a clear improvement in univariate TSFM forecasts, which remained notably less accurate than established building thermal behavior modeling approaches. While TSFMs demonstrated strong zero-shot representation capabilities, their performance still lagged behind traditional statistical methods for large datasets. Interestingly, TSFMs tend to outperform statistical models when data dynamics exhibit greater variability. Furthermore, our results demonstrate that the relative performance of models significantly depends on the evaluation metric used, reflecting their training objectives, which emphasize different priorities.

In conclusion, we believe that the effectiveness of TSFMs can be further enhanced by integrating essential metadata and accounting for confounding variables, which are critical in the context of building physics. Our future research aims to improve the applicability of TSFMs for building energy management by incorporating physics-based insights and contextual metadata into these models.

\begin{Backmatter}

\paragraph{Acknowledgments}
We thank Kang Yang for his helpful discussion and feedback on the paper.

\paragraph{Funding Statement}
This research was sponsored by Air Force Office of Scientific Research under grants \# FA95502210193 and FA95502310559, NSF under grant \# 2325956, DEVCOM ARL under Cooperative Agreement \# W911NF-17-2-0196; and, the NIH mDOT Center under award \# P41EB028242.

\paragraph{Competing Interests}
Mario Bergés and Mani Srivastava hold concurrent appointments as Amazon Scholars, and as Professors at their respective universities, but work in this paper is not associated with Amazon. Dezhi Hong is also affiliated with Amazon but work in this paper is not associated with Amazon. The views and conclusions contained in this document are those of the authors and should not be interpreted as representing the official policies, either expressed or implied, of the funding agencies. 

\paragraph{Data Availability Statement}
Except the data used in Section \ref{sec:behavior}, all data are publicly available prior to publication of this work. As part of this paper, we are publishing our implementation details and the previously-unpublished data in \url{https://github.com/nesl/TSFM_Building}.

\paragraph{Ethical Standards}
The research meets all ethical guidelines, including adherence to the legal requirements of the study country.

\paragraph{Author Contributions}
Conceptualization: O.B.M; P.Q; X.O; D.H; M.B; M.S.
Methodology: O.B.M; P.Q; L.H; DH; M.B; M.S.
Data curation: O.B.M; P.Q; L.H.
Formal analysis: O.B.M; P.Q; L.H.
Investigation: O.B.M; P.Q; X.O; DH; M.B; M.S.
Validation: O.B.M; PQ; LH.
Visualization: OBM; PQ.
Writing – original draft: O.B.M; P.Q; L.H.
Writing – review \& editing: O.B.M; P.Q; X.O; D.H; M.B; M.S.
Project administration: X.O; D.H; M.B; M.S.
Supervision: X.O; D.H; M.B; M.S.
Funding acquisition: M.B; M.S. All authors approved the final submitted draft.

%\renewcommand{\bibpreamble}{By default, this template uses \texttt{bibtex} and adopts the AMS referencing style. However, the journal you’re submitting to may require a different reference style; specify the journal you're using with the class' \texttt{journal} option --- see lines 1--19 of \emph{sample.tex} for a list of options and instructions for selecting the journal.}

% If using any of the following journal options:
%   wet, dap, dce, eds, prm, flw, jdm, psy, rsm
% then use the \printbibliography line instead of:
%\bibliography{sample-base}
\printbibliography

\end{Backmatter}

\end{document}